\documentclass[a4paper,fleqn]{cas-dc}

\usepackage[numbers]{natbib}

\usepackage{threeparttable, booktabs}
\usepackage{amsmath}
\usepackage{amsfonts}
\usepackage{amssymb}
\usepackage{bbm, bm}
\usepackage{hyperref}
\usepackage{color}
\usepackage{multirow}
\usepackage{enumitem}
\usepackage{setspace}

\def\tsc#1{\csdef{#1}{\textsc{\lowercase{#1}}\xspace}}
\tsc{WGM}
\tsc{QE}
\tsc{EP}
\tsc{PMS}
\tsc{BEC}
\tsc{DE}

\begin{document}
	\let\WriteBookmarks\relax
	\def\floatpagepagefraction{1}
	\def\textpagefraction{.001}
	\shorttitle{}
	\shortauthors{Jiahua Du et~al.}
	
	\title [mode = title]{Exploiting Review Neighbors for Contextualized Helpfulness Prediction}
	
	\author[1]{Jiahua Du}
	\cormark[1]
	\cortext[cor1]{Principal corresponding author}
	\ead{yukachan.d@gmail.com}
	\credit{Conceptualization; Data curation; Formal analysis; Methodology; Software; Validation; Visualization; Writing - original draft; Writing - review \& editing}
	\address[1]{Institute of Sustainable Industries \& Liveable Cities, Victoria University, Melbourne, VIC, Australia}
	
	\author[2]{Jia Rong}
	\cormark[2]
	\cortext[cor2]{Corresponding author}
	\ead{jiarong@acm.org}
	\credit{Methodology, Validation, Supervision}
	\address[2]{Faculty of Information Technology, Monash University, Clayton, VIC, Australia}
	
	\author[1]
	{Hua Wang}
	\credit{Supervision}
	\ead{hua.wang@vu.edu.au}
	
	\author[1]
	{Yanchun Zhang}
	\credit{Supervision}
	\ead{yanchun.zhang@vu.edu.au}
	
	\begin{abstract}
		Helpfulness prediction techniques have been widely used to identify and recommend high-quality online reviews to customers.
		Currently, the vast majority of studies assume that a review's helpfulness is self-contained.
		In practice, however, customers hardly process reviews independently given the sequential nature.
		The perceived helpfulness of a review is likely to be affected by its sequential neighbors (i.e., context), which has been largely ignored. 
		This paper proposes a new methodology to capture the missing interaction between reviews and their neighbors.
		The first end-to-end neural architecture is developed for neighbor-aware helpfulness prediction (NAP).
		For each review, NAP allows for three types of neighbor selection: its preceding, following, and surrounding neighbors.
		Four weighting schemes are designed to learn context clues from the selected neighbors.
		A review is then contextualized into the learned clues for neighbor-aware helpfulness prediction. 
		NAP is evaluated on six domains of real-world online reviews against a series of state-of-the-art baselines.
		Extensive experiments confirm the effectiveness of NAP and the influence of sequential neighbors on a current reviews.
		Further hyperparameter analysis reveals three main findings.
		(1) On average, eight neighbors treated with uneven importance are engaged for context construction.
		(2) The benefit of neighbor-aware prediction mainly results from closer neighbors.
		(3) Equally considering up to five closest neighbors of a review can usually produce a weaker but tolerable prediction result.
	\end{abstract}
	
	
	\maketitle
	
	\section{Introduction}
	
	User-generated reviews play an integral part in contemporary online shopping activities.
	A recent survey \cite{BrightLocal2017} shows that $97$\% of customers rely on online reviews to make everyday decisions.
	Moreover, $85$\% of the customers perceive the reviews as personal recommendations.
	Online reviews provide new customers with opinions and experience written by previous buyers.
	From manufactures' perspective, online reviews also help understand consumer needs and improve product quality.
	Nonetheless, online reviews are uneven in quality.
	As a product accumulates reviews, high-quality reviews may be buried by the others of random quality.
	The increasing challenge requires automatic approaches for locating helpful reviews against information overload.
	
	Helpfulness prediction aims to identify and recommend high-quality reviews to customers.
	Prior literature \cite{Ocampo2018, Hoffait2018, Charrada2016} has explored various features and models.
	One critical drawback of most existing work is the assumption that customers are unbiased and process reviews independently.
	In other words, a review's helpfulness is assumed to be self-contained.
	In practice, however, customers often read multiple reviews \cite{BrightLocal2017, Askalidis2016} before making final decisions.
	Since online reviews are sequentially displayed, how and where a review is positioned \cite{Sipos2014} can potentially affect customers' perception of helpfulness.
	In this case, the received votes of a review may not only depend on itself but also the comparison with its the surrounding reviews.
	
	\begin{table}[width=.95\linewidth,cols=3,pos=t]
		\centering
		\caption{The perceived helpfulness of a review (\#3) can be affected by its neighbors (\#1 and \#2).}
		\label{tab:context:example}
		\begin{tabular*}{\tblwidth}{@{} llp{0.35\textwidth}@{} }
			\toprule
			& \multicolumn{2}{l}{Review} \\ \midrule
			\multirow{3}{*}{(a)} & \#1 & This headphone is soooo cool! \\ 
			& \#2 & Best headphone in my life. I would definitely recommend it!!! \\
			& \#3 & The headphone has a fashionable appearance and the sound quality is excellent. I am surprised that it's even waterproofed. \\
			\hline
			\multirow{3}{*}{(b)} & \#1 & You can't find any headsets better than this. \\ 
			\multicolumn{1}{c}{} & \#2 & Cheap price with good quality. \\ 
			& \#3 & The advertisement says the headphone can last for $10$ hours with full battery. Well, obviously it doesn't. \\ \bottomrule
		\end{tabular*}
	\end{table}
	
	Table~\ref{tab:context:example} illustrates the idea with two toy examples.
	Assuming that customers read the reviews in order.
	In example~(a), review $\#1$ and $\#2$ set a positive impression of a headphone product.
	Review $\#3$ shares a similar and yet more detailed opinion, which reinforces the impression.
	Within the context of review $\#1$ and $\#2$, review $\#3$ is seemingly more convincing and likely to receive higher helpfulness than by itself.
	Example~(b) shows another situation where review $\#3$ provides new information (i.e., defects of the headphone) that differs from review $\#1$ and $\#2$.
	In this case, review $\#3$ can be more helpful due to the new information, or less helpful due to contrasting the existing impression.
	Both examples indicate that the perceived helpfulness of a review is not always self-contained nor independent, and the influence of a review's neighbors should be taken into account.
	
	Different from the vast majority of prior research, this work hypothesizes that the helpfulness of a review not only depends on itself but also its neighbors. 
	A deep neural architecture is proposed for Neighbor-Aware helpfulness Prediction (NAP).
	NAP first learns representations for individual reviews.
	For each review, three intuitive types of review neighbors are considered: (1) preceding reviews, (2) following reviews, and (3) surroundings reviews.
	Four weighting schemes are then explored to construct context from the neighbor representations.
	During helpfulness modeling, the interaction between a review and its neighbors is captured by aggregating the contextual clues.
	
	Note that the terms ``context'' and ``neighbor'' have been used considerably differently \cite{Ocampo2018} in helpfulness prediction and pertinent fields.
	In most cases, context indicates information extracted from the same review as opposed to content, namely, review texts.
	Such information includes product metadata \cite{Ghose2011}, reviewer characteristics \cite{Huang2015}, and reviewer historical voting data \cite{Ghose2011,Huang2015,Lu2010}.
	Still, reviews are treated independently and no review interaction is captured.
	Context can also suggest information beyond individual reviews. 
	In \cite{Lu2010,Moghaddam2012,Tang2013}, reviews are interacted via user idiosyncrasies and rater-reviewer social connections.
	Under this setting, users with similar preferences \cite{Hao2011,Sarwar2001} are occasionally referred to as neighbors.
	In \cite{Sipos2014}, neighbors are defined as surrounding reviews of a given review.
	While the former type of neighbors have been broadly researched, the influence of the latter remains understudied.
	
	This work targets neighbor-aware helpfulness prediction.
	Specifically, the helpfulness of each review is contextualized into its neighbors.
	Similar to \cite{Sipos2014}, neighbors are clarified as adjacent reviews of a given review in a review sequence displayed to customers.
	The terms ``neighbor-aware'' and ``contextualized'' are henceforth used interchangeably.
	On the other hand, methods that only depend on information within individual reviews are called independent helpfulness prediction. 
	More details of independent and contextualized helpfulness modeling will be discussed in Section~\ref{sec:context:lr}.
	
	To the best of our knowledge, this work offers the following contributions:
	\begin{enumerate}
		\item \textbf{End-to-end neighbor-aware helpfulness}: This work is one of the pioneer studies considering the interaction between a review and its neighbors when modeling helpfulness.
		Previous work majorly interacts reviews from a global perspective, using the whole review collection as context.
		This work instead aims at the local interaction (i.e., neighbors) among reviews.
		NAP also provides the first end-to-end solution for contextualized helpfulness modeling.
		\item \textbf{Comprehensive contextual settings}: NAP allows for three neighbor selection and four weighting schemes for context construction. 
		To ensure the flexibility of neighbor utilization, the four weighting schemes (each with increasing learning parameters) construct contextual information from a various number of preceding, following, and surrounding neighbors.
		
		\item \textbf{Extensive evaluation and analysis}: 
		A series of experiments are conducted to evaluate the effectiveness of NAP.
		Hyperparameter studies are further analyzed investigate model sensitivity to discuss the trade-off between model complexity and performance.
		Qualitative analysis  provides visualization and case studies for better understanding the model interpretation. 
		Experimental results show NAP is effective in neighbor-aware helpfulness prediction and offer insights into utilizing neighbors for the task.
	\end{enumerate}
	
	The remaining of the paper is organized as follows.
	Section \ref{sec:context:lr} surveys existing studies on independent and context-aware helpfulness prediction.
	Section \ref{sec:context:method} formalizes the problem of neighbor-aware helpfulness prediction and presents the NAP framework.
	Section \ref{sec:context:exp} describes experiment settings for evaluating NAP against a series of baselines.
	Section \ref{sec:context:results} demonstrates the effectiveness of NAP, performs sensitivity analysis on contextual settings, and provides qualitative analysis on the trained models.
	Section \ref{sec:context:con} summarizes findings and discusses future research directions.
	
	\section{Related Work}
	\label{sec:context:lr}
	
	Helpfulness prediction can either be approached in an independent or contextualized manner.
	The former (as most studies did) assumes that the helpfulness of a review is self-contained.
	The latter adopted by more recent studies considers helpfulness as an interactive function of a review and its counterparts.
	The following subsections survey literature on the two categories of helpfulness prediction and discuss social influence on helpfulness perception.
	
	\subsection{Independent Helpfulness Prediction}
	
	The vast majority of existing work predicts a review's helpfulness merely using information contained in itself. 
	In the past decade, a large body of hand-crafted features \cite{Ocampo2018, Hoffait2018,Charrada2016,Du2019Selection} have been carefully curated to represent the helpfulness of a review,
	including review text \cite{Du2019Selection,MalikHussain2017}, review metadata \cite{Willemsen2011,Kuan2015}, and reviewer characteristics \cite{Cheng2015,Hu2016}.
	Once chosen, the features are concatenated to represent a review and then fed into traditional machine learning algorithms for helpfulness prediction.
	Such methodology has the merit of easy implementation and clear interpretation due to the feature engineering nature.
	However, preparing effective features requires domain-specific expert knowledge, which is laborious.
	
	Recent studies approach the task via deep learning techniques.
	With neural architectures, the latent representations encoding helpfulness are learned automatically, bypassing the tedious feature engineering \cite{Ocampo2018} process.
	Currently, models developed upon convolutional neural networks (CNNs) \cite{Kim2014CNN, Kim2016} and recurrent neural networks \cite{Lipton2015RNNcritical} such as long short-term memory networks (LSTMs) \cite{Hochreiter1997LSTM} and gated recurrent units (GRUs) \cite{Cho2014GRU} have shown to be feasible for helpfulness feature learning.
	
	Saumya et al. \cite{Saumya2019} employ a two-layer CNN to encode review texts.
	Chen et al. \cite{Chen2018} consider helpfulness modeling as a cross-domain task.
	To alleviate the out-of-vocabulary issue, subword information is integrated into word-level review representations.
	Three CNNs are separately built on top of the embeddings to transfer knowledge: one summarizes common knowledge shared across domains; the other two learn domain-specific knowledge.
	In another work, Chen et al.  \cite{Chen2019gcnn} extends the framework to conduct multi-domain helpfulness prediction.
	In addition to subword information, word embeddings are further enhanced with the distribution of product aspects \cite{Yang2016} mentioned in reviews.
	In addition, gating mechanisms are adopted to learn multi-granularity text features that identify word importance in reviews.
	
	Qu et al. \cite{Qu2018} propose two CNN variants to combine review texts and star ratings for helpfulness prediction.
	The first method attaches raw star ratings as an extra dimension to the learned content representations.
	The second method treats each star rating as a part (the last word) of a review.
	Star ratings are embedded and then attached to the word embedding matrix for content representation learning.
	Although star embeddings enable larger encoding capacity, the current integration method largely restricts rating information from interacting with review content.
	Du et al. \cite{Du2019ECRI} cope with the issue by separating the encoding of rating embeddings from that of review content.
	To ensure the direct influence of star ratings on review texts, star embeddings are aligned to and then interacted with the convoluted content embeddings.
	
	Fan et al. \cite{Fan2018,Fan2019} integrate rating information by formulating helpfulness prediction as a multi-task learning problem.
	In \cite{Fan2018}, an attention-based CNN is employed to encode review texts.
	In \cite{Fan2019}, the authors model into helpfulness the semantic closeness of review texts reflecting on characteristics mentioned in the targeted product title.
	Two sets of bidirectional LSTMs are first used to learn separate representations for review texts and the product title.
	The closeness is then measured via attention mechanisms, which are used to reinforce review representations.
	The learned representations in both cases are then used to predict the helpfulness of a review and the accompanying star rating  simultaneously.
	
	Ma et al. \cite{Ma2018} investigate the extent to which photos posted along with reviews influence the perceived helpfulness in the hotel industry.
	To this end, text representations are learned by LSTMs, whereas image representations is obtained via a pre-trained $152$-layer deep residual network \cite{He2016Deep}.
	Both learned representations are then concatenated and fed into another LSTM to predict review helpfulness.
	
	The independent assumption helps simplify the process of data preparation and model construction.
	As will be discussed, however, human helpfulness perception is more complicated and involves a variety of social biases.
	As such, the assumption may lead to unreliable and problematic prediction in practice.
	This work instead hypothesizes that a review's helpfulness depends on both itself and the context it is fit into.
	More specifically, the context of a review is referred to as information learned from its spatial neighbors.
	
	\subsection{Social Influence on Helpfulness Perception}
	
	Social influence \cite{Choi2018Repurchase, Momeni2015Survey} has been proven to be a key part in decision making through extensive experiments \cite{Salganik2006, Muchnik2013, Banerjee1992, Grabowicz2015, Mercier2019} in psychology, economics, sociology, and human behavior analysis.
	The core idea of social influence is that one's decision can be affected by the presence and behavior of others \cite{Choi2018Repurchase, Momeni2015Survey, Danescu2009, Liu2015Agent, Berger2016Presentation}.
	Such influence also takes effect among strangers \cite{Cialdini2004social} and in online environments \cite{Xie2019Understanding, Liu2016Rating, Cosley2003}.
	In the context of helpfulness perception, decision making refers to customers perusing online reviews and then determining the extent to which the reviews are helpful.
	Currently, the perception process is subjective varying from customers.
	Thus, the task is vulnerable to social influence.
	
	Many existing studies \cite{Asch1946forming, nakayama2012exploratory, Kapoor2009, Qiu2011Effects} attribute the social influence on helpfulness perception to the sequential nature of online reviews.
	Since reviews are sequentially displayed, how a review is positioned and presented \cite{Danescu2009} to customers can affect its perceived helpfulness.
	Qiu et al. \cite{Qiu2011Effects} confirm the presentation order of positive and negative reviews can influence the cognitive outcomes of readers.
	A line of experimental studies \cite{Sikora2012Kalman,Wan2012Biased,Page2010last,Shrihari2012,Wendy2011} conclude that customers are biased by past reviews when processing subsequent ones.
	In \cite{Sipos2014}, Sipo et al. observe helpfulness voting being used as adjustment to ``correct'' reviews that customers believe should have a lower/higher ranking in the sequence.
	In consequence,  helpfulness evaluation rarely takes place independently.
	The findings above have been adopted in star rating prediction \cite{Liu2016Rating, Guo2016Understanding, Wang2014Quantifying}, yet little is known how review order influences review helpfulness perception.
	
	Recent studies further reveal the role of review order in helpfulness perception.
	One plausible explanation is the confirmation bias.
	Customers usually have their own understanding and thoughts (initial beliefs) towards products before searching.
	In this case, the goal of reading reviews is to gain further confirmation to support the preset expectation.
	When encountering a review that deviates from the expectation, customers may perceive the review as less helpful since the expressed opinions violates their initial belief.
	As stated in \cite{Yin2015}, more certain (uncertain) initial beliefs may lead to more (less) pronounced confirmation bias.
	
	Another similar explanation is the anchoring effect.
	Unlike the confirmation bias where customers hold their own initial beliefs, the first impression \cite{Rabin1999first} is formed during review perusal.
	According to Daomeng et al. \cite{Daomeng2019}, customers establish a reference frame \cite{Zhang2020Effect} to evaluate their personal voting behavior.
	A relative majority opinion is learned from past reviews and compared with subsequent reviews.
	The majority opinion sets the initial beliefs (i.e., anchor), whereas the subsequent ones serve as new opinions.
	When comparing the two types of opinions \cite{Zhang2017Contrast, Quaschning2015, Lopez2016most, Danescu2009} (in terms of text informativeness, valence, etc.), the resulting (in)consistency \cite{Walther2012Congruity} can affect customers' perception.
	Zhang et al. \cite{Zhang2017Contrast} summarize three evaluation patterns for the (in)congruent opinions using the assimilation and contrast theories.
	
	Last but not least, review helpfulness can be explained using the information theory---whether a review provides new information. 
	Jorge et al. \cite{Jorge2019Entropy} argue that if later reviews provide little or no new information apart from what has been described in early ones, themselves may be less helpful regardless of quality.
	A similar view is addressed in \cite{Tsur2009Revrank}.
	If words in a review are partly shared by other reviews, the review is to a certain extent predictable based on previous ones.
	Hence, the review is of lower uncertainty and expected to be less helpful to a reader.
	
	\subsection{Contextualized Helpfulness Prediction}
	
	Few studies have attempted to integrate information beyond individual reviews into helpfulness modeling.
	It is worth noting again that the term ``contextualized'' investigates the interaction between a review and its surrounding neighbors rather than that in \cite{Lu2010,Moghaddam2012,Tang2013} modeling user idiosyncrasies and rater-reviewer social connections.
	
	Zhou et al. \cite{Zhou2017Order} form an order variable to assess the impact of sequential dynamics.
	The authors follow \cite{Silva2012} and first sort dynamically-ranked reviews by their time stamps.
	The variable then records the position of individual reviews, where those posted on the same day share an identical position.
	The extracted order is then used as one of the variables to construct prediction models.
	The same review orders are also adopted by \cite{Zhou2017Order, Zhou2019roles, Fresneda2019, Jorge2019Entropy}.
	Alzate et al. \cite{Alzate2018Exploring} introduce three types of review orders into feature engineering: reviews in a sequence that are ranked by (i) newest review, (ii) most helpful review, and (iii) highest rating review.
	The authors further normalize the orders into probability variables to smooth the model interpretation.
	
	Lu et al. \cite{Lu2010} measure the review conformity by comparing the word distribution of a review with that of the others.
	The authors first vectorize reviews via a unigram language model.
	The overall opinion is set as the average of all review vectors related to the same item.
	The conformity results from the Kullback--Leibler divergence between a review representation and the overall opinion.
	
	Hong et al. \cite{Hong2012} measure the sentiment divergence of a review from the mainstream opinion of an item. 
	The polarity (i.e., positive, neutral, negative) of each review is first identified based on the percentage of positive and negative words in a review.
	The mainstream opinion belongs to the valence that shared by the majority of reviews of the same item.
	The divergence between a review and the mainstream opinion is defined as their valence difference.
	
	In \cite{Jorge2019Entropy}, the authors measure the incremental information entropy of each review.
	The entropy is defined as the number of new words in a current review beyond that have been mentioned in the manufacturer-provided product description and in its previous reviews.
	
	These approaches mainly suffer from three drawbacks.
	First, many platforms constantly update review orders as helpfulness voting evolves.
	Apparently, one single snapshot of reviews cannot reflect the ranking dynamics \cite{Eryarsoy2014} over time. 
	Therefore, most of the studies are not modeling the true order information.
	A possible solution to cope with the issue is to obtain multiple snapshots \cite{Alzate2018Exploring, Lee2016Exchangeability, Sipos2014} of the same set of reviews, but the task is time-consuming and limited to small datasets.
	Deciding the time granularity is also difficult.
	Second, customers are assumed to be aware of the whole review collection of an item (i.e., global context) when determining a review's helpfulness.
	As discussed, customers only have limited patience for few reviews, and thus the assumption is hardly possible in reality.
	Third, most of the methods focus on peripheral cues \cite{Mousavizadeh2015} of reviews for helpfulness modeling.
	Features derived from review texts, which arguably contain the richest information, remain underdeveloped.
	
	This work extends and differs from existing literature as follows. 
	(1) A novel dataset containing six domains of online reviews is created for experimentation.
	The dataset is advantageous since reviews regularly uploaded by new customers are consistently ranked in reverse chronological order.
	(2) Deep neural techniques are employed to offer an end-to-end solution that directly learns contextualized features from review texts.
	(3) Local context is adopted in place of the global counterpart and constructed in a more flexible and comprehensive manner.
	
	\section{Neighbor-aware Prediction Networks}
	\label{sec:context:method}
	
	The research problem of neighbor-aware helpfulness prediction is formulated as a binary text classification task.
	Without loss of generality, let $\mathbf{S}=(S_1, S_2, \ldots, S_N)$ be an ordered list of $N$ reviews and $\mathbf{y}=(y_1, y_2, ..., y_N)$ the corresponding helpfulness labels, where $y=1$ is helpful and $y=0$ unhelpful. 
	Most existing studies oversimplify helpfulness prediction of a review $S_i (i \in [1,N])$ using the independent assumption $P(y \mid S_i; \theta)$, where $\theta$ are model parameters. 
	Such approaches are henceforth called \textit{independent} helpfulness prediction.
	NAP instead associates $S_i$ with a context $\mathbf{T}_i$ composing reviews selected from its neighbors.
	The goal of NAP is to predict the probability of $S_i$ being helpful $P(y \mid S_i, \mathbf{T_i}; \theta)$, and thus \textit{contextualized} review helpfulness prediction. 
	
\begin{figure*}[t]
		\centering
		\includegraphics[width=1\linewidth]{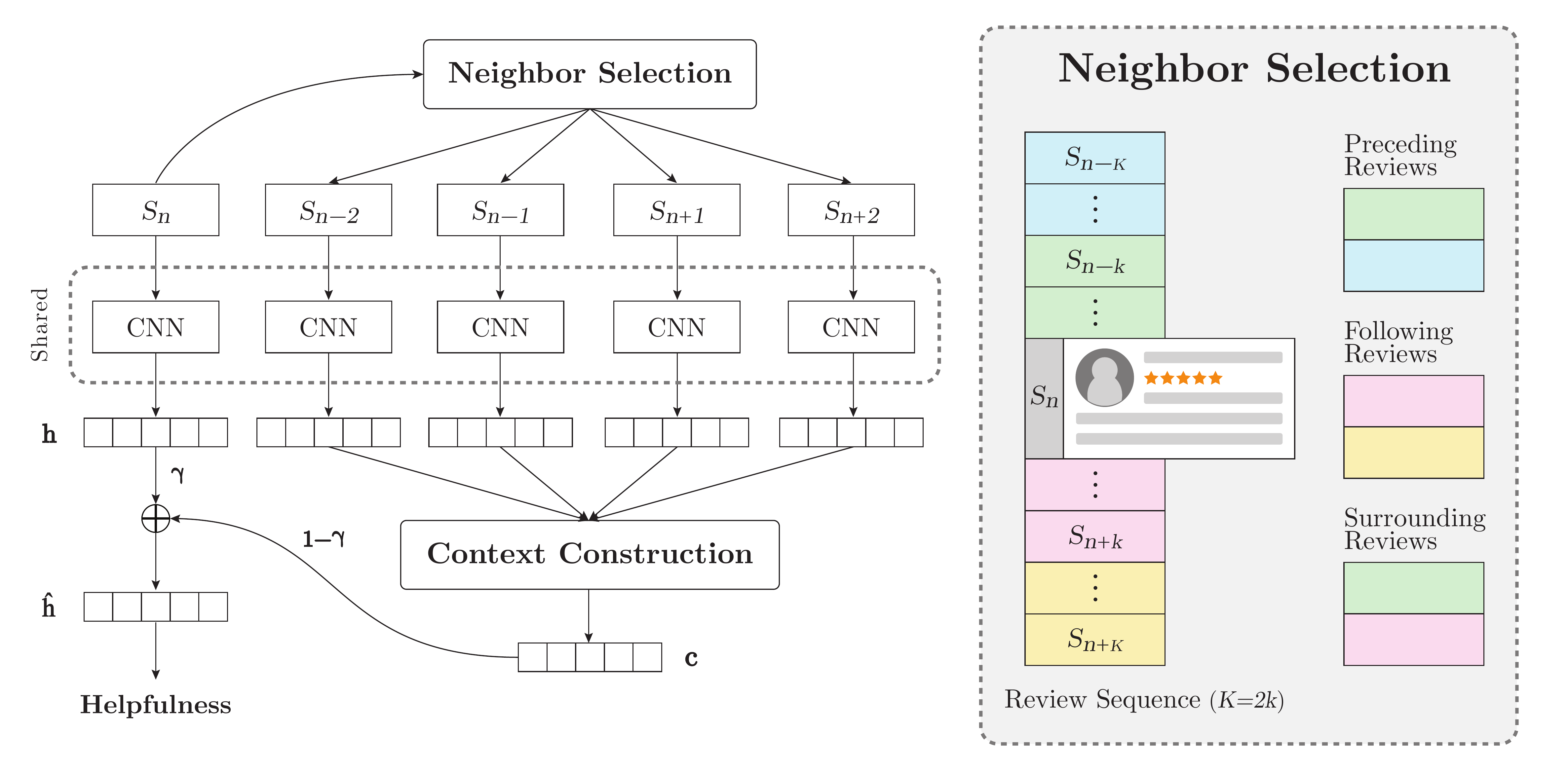}
		\caption[The NAP architecture.]{The NAP architecture.
			As an example, $K=4$ surrounding reviews are selected as neighbors to construct the context of the current review $S_n$. 
		}
		\label{fig:context:model}
	\end{figure*}

	This section presents NAP, an end-to-end deep neural architecture for the task.
	As illustrated in Figure~\ref{fig:context:model},
	NAP consists of three learning phases. 
	The review encoding phase transforms each review $S$ into an embedding $\textbf{h}$.
	The context construction phase combines the embeddings of the associated context $\mathbf{T}_i$ into a context embedding $\textbf{c}$.
	Finally, $\textbf{h}$ and $\textbf{c}$ are aggregated to obtained the neighbor-aware representation of $S$ used for helpfulness prediction.
	The following subsections detail each model component of NAP.
	
	\subsection{Review Text Encoding}
	
	Let each review $S=(x_1, x_2, \ldots, x_n)$ be a sequence of $n$ words.
	The vocabulary $V$ is constructed via indexing all unique words in $\mathbf{S}$. 
	Given an embedding lookup table $\mathbf{E} \in \mathbb{R}^{|V| \times d}$, each word $x \in V$ is associated with a $d$-dimensional word vector $\mathbf{e}_{x} \in \mathbf{E}$.
	Specifically, $x$ is encoded using the one-hot encoding scheme into $\mathbf{x} \in \mathbb{R}^{|V|}$ to select the corresponding word vector $\mathbf{e}_{x}$.
	As a result, $S$ can be represented by an embedding matrix $\mathbf{X} \in \mathbb{R}^{n \times d}$: 
	\begin{align}
		\mathbf{e}_{x} = & \mathbf{E}^\top \mathbf{x}, \\
		\mathbf{X} = & [\mathbf{e}_{x_1}, \mathbf{e}_{x_2}, \ldots, \mathbf{e}_{x_n}].
	\end{align}
	
	The CNN framework proposed by Kim \cite{Kim2014CNN} is used to further encode the semantic meaning of review texts.
	Note that the goal of this paper is neighbor-aware helpfulness prediction as a proof of concept.
	The focus is incorporating review neighbors as context information instead of complex model construction.
	Therefore, the vanilla CNN framework is chosen to control the total number of training parameters.
	To learn more sophisticated review representations one could use more advanced CNN frameworks \cite{Du2019ECRI} and learn adaptive word- and character-level \cite{Chen2018, Chen2019gcnn} embeddings.
	
	NAP employs $m$ kernels for convolution. 
	Each kernel is applied to a sliding window of $l$ words over $\mathbf{X}$ to produce new features.
	The convoluted features are then activated using Exponential Linear Unit (ELU) \cite{clevert2015fast} function:
	\begin{align}
		\mathbf{H} = \text{ELU}(\mathbf{X} \ast \mathbf{W}_c + \mathbf{b}_c
		),
	\end{align}
	where the kernels $\textbf{W}_c \in \mathbb{R}^{l \times d \times m}$ and biases $\textbf{b}_c \in \mathbb{R}^{m}$ are parameters to be estimated.
	
	The embedding of individual reviews $\mathbf{h}$ is then obtained via column-wise max pooling \cite{collobert2011natural} over the feature maps:
	\begin{align}
		\mathbf{h} = \max (\mathbf{H}).
	\end{align}
	
	\subsection{Neighbor-aware Context Construction}
	
	NAP constructs the context of a review from its neighbors.
	Specifically, each review $S_i \in \mathbf{S}$ in the sequence is associated with a context $\mathbf{T}_i$ of $K=2k, k \in \mathbb{N}^{+}$ reviews selected from its neighbors $\{S_j \mid j \in [i-2k, i+2k], j \neq i\}$.
	Three neighbor selection schemes are explored for context construction.
	\begin{equation}
		\mathbf{T}_i=
		\begin{cases}
			(S_j)_{j=i-2k}^{i-1}, & \text{$K$ preceding reviews,} \\
			(S_j)_{j=i+1}^{i+2k}, & \text{$K$ following reviews,} \\
			(S_j)_{j=i+k}^{i-k} \setminus S_i, & \text{$K$ surrounding reviews.}
		\end{cases}
		\label{eq:context:neighbors}
	\end{equation}
	
	The context $\mathbf{T}_i$ is regarded as reviews a user has previously read prior to the current one $S_i$.
	NAP accepts both preceding and following reviews as context because the review order in $\mathbf{S}$ does not necessarily reflect the reading order.
	In addition, users can vote the helpfulness of a review straight after the perusal or after reading other reviews.
	It can be seen that the current review $S_i$, by definition, can also be part of the context of other reviews.
	
	To learn the context of a review $S$, the selected neighbors are mapped into embeddings and further stacked into an embedding matrix $\mathbf{C} \in \mathbb{R}^{K*m}$.
	The context embedding, denoted by $\mathbf{c} \in \mathbb{R}^{m}$, is calculated by transforming $\mathbf{C}$ via a weighting scheme $f:\mathbb{R}^{K*m} \to \mathbb{R}^{m}$, $\mathbf{c} = f(\mathbf{C})$.
	When $K>1$, $f$ merges the $K$ neighbor embeddings, which 
	imitates customers learning the first impression $\mathbf{c}$ from past reviews $\mathbf{C}$.
	The weights indicate the influence of individual reviews perceived by customers.
	When $K=1$, $f$ is an identity map since one neighbor contains all information and no combination is required.
	
	NAP introduces four weighting schemes to merge $K$ neighbor embeddings.
	Each scheme is a special case of its following one, with increasing flexibility in parameter learning. 
	
	\begin{enumerate}
		\item \textbf{Average (AVG)}
		The first weighting scheme borrows the idea from the neural bag-of-words model \cite{mikolov2013distributed}.
		In the model, a sentence embedding results from the centroid of its constituent word counterparts, which can be thought of as a summary of the sentence.
		This simple model has been used in many natural language processing tasks \cite{Arora2017asimple,Wieting2015towards,Iyyer2015Deep} and proven robust and effective.
		Here, the context (analogous to a sentence) embedding is represented as the bag-of-reviews representation of the $K$ neighbors (analogous to words).
		\begin{align}
			\textbf{c} = \frac{1}{K} \sum_{i=1}^K \textbf{C}_i.
			\label{eq:context:avg}
		\end{align}
		
		The AVG scheme requires no parameters for context construction.
		The identical weights show equal importance of individual reviews when customers' composing their first impression towards a product.
		
		\item \textbf{Weighted Average (WAVG)}
		The second weighting scheme extends AVG.
		In reality, user-generated reviews are uneven in quality, text valence, and sentiment intensity.
		Assigning separate importance for individual reviews
		provides higher flexibility in context construction.
		As such, the fixed weights in Equation \eqref{eq:context:avg} are replaced by parameters learned via an attention mechanism \cite{Raffel2015feed}, which employs a query vector $\textbf{u}_a \in \mathbb{R}^{m}$ as the learnable function:
		\begin{align}
			z_i & = \text{tanh}(\textbf{u}_a^{\top}\textbf{C}_i),
			\\
			\alpha_i & = \frac{\exp(z_i)}{\sum^K_{j=1}\exp(z_j)},
			\\
			\textbf{c} & = \sum^K_{i=1} \alpha_i \textbf{C}_i,
		\end{align}
		The context embedding is then obtained from the weighted average of the $K$ review embeddings. 
		
		\item \textbf{Feature Regression (FR)}
		The third weighting scheme further extends WAVG.
		Each dimension of a review embedding suggests a certain type of latent review characteristic.
		During perusal, different characteristics may attract various interests.
		Thus, combining review embeddings on a dimension level enables more flexibility in utilizing the relationship across features.
		The weights are computed using a similar attention mechanism. 
		Specifically, the context matrix $\mathbf{C}$ is first transformed into $\mathbf{Z} \in \mathbb{R}^{K*m}$ via another matrix of the same shape, followed by column-wise softmax normalization.
		\begin{align}
			\mathbf{Z} & = \text{tanh}(\mathbf{W}_b \otimes \mathbf{C}),
			\label{eq:context:fr1}
			\\
			\beta_{ij} & = \frac{\exp(\mathbf{Z}_{ij})}{\sum^K_{k=1}\exp(\mathbf{Z}_{kj})},
			\label{eq:context:fr2}
			\\
			c_j & = \sum^K_{k=1} \beta_{kj} \mathbf{C}_{kj},
			\label{eq:context:fr3}
		\end{align}
		where $\mathbf{W}_b \in \mathbb{R}^{K*m}$ are learned parameters and $\otimes$ the Hadamard product.
		The $j$-th dimension $c_j$ is then the weighted average of the same context matrix column $(\mathbf{C}_{kj})_{k=1}^K$.
		The result of $c_j$ can also be thought of as conducting linear feature regression on $(\mathbf{C}_{kj})_{k=1}^K$.
		
		\item \textbf{Spatial Feature Regression (SFR)}
		The fourth weighting scheme considers the interaction among neighbors.
		Since reviews are sequentially displayed, neighbors closer to the target review are more likely to attract higher reading priority.
		In addition, neighbors being read earlier may influence those later.
		To capture such influence, information of closer neighbors is shared with farther ones such that:
		\begin{equation}
			\hat{\mathbf{C}}_i =
			\begin{cases}
				\sum^{K}_{k=i} \mathbf{C}_k, & \text{Preceding reviews,} \\
				\sum^{i}_{k=1} \mathbf{C}_k, & \text{Following reviews.} \\
			\end{cases}
		\end{equation}
		
		As for surrounding reviews, the left half and right half are regarded as preceding and following reviews, respectively. 
		The enhanced context matrix $\hat{\textbf{C}}$ is then passed to Equations \eqref{eq:context:fr1}--\eqref{eq:context:fr3} in place of $\textbf{C}$ for context construction.
	\end{enumerate}
	
	\subsection{Contextualized Helpfulness Prediction}
	
	Finally, NAP contextualizes a review within its neighbors by aggregating the embedding of a review $\mathbf{h}$ and that of its neighbors (i.e., context) $\mathbf{c}$ via linear combination:
	\begin{align}
		\hat{\mathbf{h}} = \gamma \mathbf{h} + (1-\gamma) \mathbf{c}.
		\label{eq:context:combine}
	\end{align}
	Here, $\mathbf{c}$ learns the relative majority opinion \cite{Daomeng2019} that can be thought of as a user's initial belief towards an item, whereas $\mathbf{h}$ serves as a new opinion.
	The contextualization thus learns the interaction between the initial belief and new opinion. 
	The combination factor $\gamma \in [0,1]$ controls the influence of neighbors on the current review.
	Note that setting $\gamma=1$ stops the influence of neighbors.
	In this case, the helpfulness information of a review is self-contained, and thus called independent helpfulness prediction.
	When $\gamma=0$, a review's helpfulness relies exclusively on its context.
	
	The neighbor-aware representation $\hat{\textbf{h}}$ is then forwarded into a logistic regression layer to predict the helpfulness of the current review.
	\begin{align}
		\hat{y} = \sigma(\mathbf{W}_o^{\top}\hat{\mathbf{h}} + b_o).
	\end{align}
	
	NAP is trained via cross entropy minimization over $M$ samples. 
	The regularization on the CNN filters with weight decay $\lambda$ is added to reduce the overfitting of text encoding.
	\begin{align}
		\mathcal{L} = - \frac{1}{M} \Big[
		\mathbf{y}^\top \log(\hat{\mathbf{y}}) + 
		(1-\mathbf{y})^\top \log(1-\hat{\mathbf{y}})
		\Big] + 
		\frac{\lambda}{2} \lVert \mathbf{W}_c \rVert^2,
	\end{align}
	where $\hat{\textbf{y}}$ and $\textbf{y}$ are the predicted and actual helpfulness labels respectively. 
	
	\section{Experiment Settings}
	\label{sec:context:exp}
	
	NAP is evaluated and benchmarked against a series of baselines via extensive experiments.
	Section~\ref{sec:context:datasets} describes in detail the datasets used throughout the experiments, including data collection and pre-processing.
	Section~\ref{sec:context:baselines} describes the baselines using both traditional machine learning algorithms and deep learning architectures are described for performance comparison.
	Section~\ref{sec:context:hyper} presents hyperparameters for training NAP and the baseline models.
	
	\subsection{Datasets}
	\label{sec:context:datasets}
	
	One critical challenge of neighbor-aware helpfulness prediction is data preparation, which requires both a review and its neighbors.
	Currently, many platforms dynamically rank reviews based on a set of criteria.
	Such mechanisms change the neighbors of a review as helpfulness voting evolves.
	As a result, a review's neighbors at the time of data collection only reflect a single snapshot but not its previous dynamics.
	One could collect multiple snapshots of the reviews through periodically tracking their ranking statistics, but the collection is expensive, time-consuming, and difficult in deciding time granularity.
	
	This work opts for an alternative option to prepare eligible online reviews.
	Despite that many online platforms adopt dynamic ranking algorithms, several inherently rank reviews in reverse chronological order to provide customers with the latest user feedback of products/services.
	As for the latter, the static and consistent review order over time ideally compensates the necessity of multiple snapshots of reviews.
	In particular, two popular platforms meeting such criteria are considered:  SiteJabber\footnote{\url{https://www.sitejabber.com/}} and ConsumerAffairs\footnote{\url{https://www.consumeraffairs.com/}}.
	Both platforms offer a wide range of categories of user-generated reviews regarding products, retailers, and companies, with SiteJabber focusing more on websites and online businesses.
	It is worth noting that although evaluated on chronologically-ordered reviews, NAP is also applicable to reviews with ranking dynamics provided that multiple snapshots are given.
	
\begin{figure}[htbp]
		\centering
		\includegraphics[width=1\linewidth]{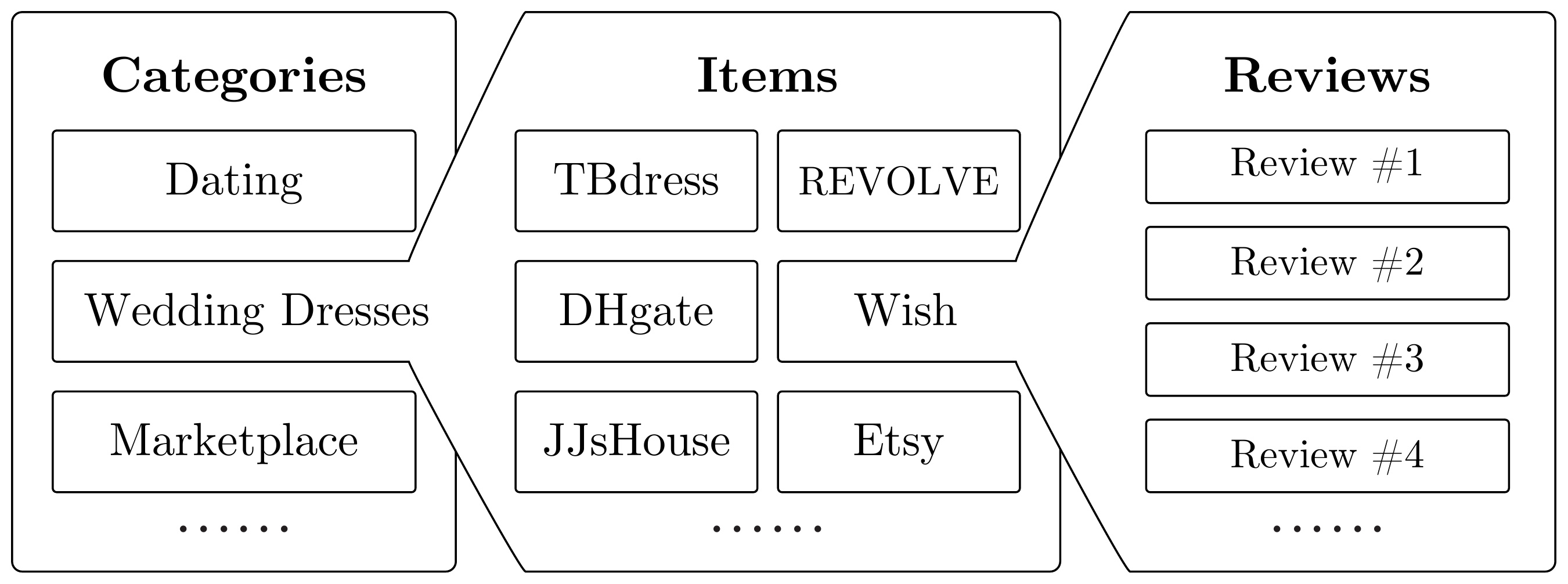}
		\caption[The hierarchy of the collected SiteJabber reviews.]{The hierarchy of the collected SiteJabber reviews. ConsumerAffairs shares the same organization.}
		\label{fig:context:dataset}
	\end{figure}

	Python scripts are compiled to crawl, extract, and store reviews from the two platforms.
	A total of $169,126$ reviews posted prior to 29 April, 2019.
	The raw SiteJabber dataset consists of $60,426$ reviews collected from three categories (i.e., Marketplace, Wedding Dresses, and Dating), whereas the ConsumerAffairs dataset originally contains $108,700$ reviews collected from the Car Insurance, Travel Agencies, and Mortgages categories.
	As shown in Figure~\ref{fig:context:dataset}, each category (domain) of a website contains a list of reviewed items and each item consists of a list of reviews.
	Table~\ref{tab:context:sample} presents two review samples for each website, along with the accompanying attributes.
	For simplicity, the six domains are called D1, D2, and so on.
	
	\begin{table*}[width=.9\textwidth,cols=2,pos=h]
		\centering
		\caption{Example SiteJabber (top) and ConsumerAffairs (bottom) review composition.}
		\label{tab:context:sample}
		\begin{tabular*}{\tblwidth}{@{} p{0.2\textwidth}p{0.61\textwidth}@{} }
			\toprule
			Attribute & Value \\
			\midrule
			Reviewer Name & David W. \\
			Total number of posts by the reviewer & 8 \\
			Total number of votes received by the reviewer & 22 \\
			Review Date & Saturday, 7 April 2018 \\
			Number of helpful votes & 10 \\
			Star Rating & 1 \\
			Review Title & They refused my order, didn't communicate or return my money \\
			Review Text & I placed an order for around \$200, The order went through and they took my money.after a little while I received an email that they put a hold on my order and the only way to have the order go through was to send them front and back pictures of my credit card and my passport. I refused but [...] \\ 
			\midrule
			Attribute & Value \\
			\midrule
			Reviewer Name & Justin \\
			Reviewer Location & Heflin, Alabama \\
			Verified Buyer & Yes \\
			Verified Reviewer & Yes \\
			Review Date & Sunday, 9 April 2017 \\
			Number of helpful votes & 8 \\
			Rating & 5 \\
			Review Text & The home loan process at Vanderbilt Mortgage was very easy going. It was also pretty fast. From the time that I went house shopping to the time that I was in my house, it took me about a month. Also, everyone I spoke to throughout the process was very informative, helpful, friendly and courteous. [...] \\
			\bottomrule
		\end{tabular*}
	\end{table*}
	
	The following pre-processing steps are applied to the raw reviews to improve data quality.
	(1) To ensure that reviews can have adequate neighbors for context assembly, only reviewed items with $\geq 100$ remaining reviews are considered.
	(2) Each review is lowercased and tokenized in to a sequence of words, followed by minimum stopword removal that eliminates articles (i.e., a, an, and the) from the reviews. 
	(3) Following \cite{Du2019ECRI}, only the most frequent $30$k terms are kept as vocabulary to reduce the execution cost during model training.
	(4) Early-posted reviews tend to receive disproportionately higher number of votes \cite{Wan2015Matthew, Wan2013Matthew} over later ones.
	To cope with the bias, reviews posted in early months that have less than $15$ reviews for the same item are removed.
	(5) 
	Similarly, 
	reviews posted recently are removed due to insufficient exposure time for voting.
	It is worth noting that reviews with few votes are usually filtered out \cite{Gobinda2018, Zhang2012} to learn more robust models. 
	Since review order is an importance factor for correctly training NAP, this work does not perform any further removal of reviews based on the number of votes.
	
	The pre-processed reviews are then labeled and split.
	Review labels are determined upon existing human assessment, namely, helpfulness votes.
	Following \cite{Ma2018}, a review is labelled as helpful if it receives at least two votes and unhelpful otherwise.
	For each domain, the constituent reviews in a reviewed item are first partitioned into three sets, using $80\%$, $10\%$, and $10\%$ of the data respectively for training, validation, and testing.
	In particular, chronological split \cite{ShengTun2019, Maroun2016, Lu2010, Moghaddam2012, Zhang2011Information} is applied over randomization to preserve the review order information.
	
	After dataset partition, review words that are numeric values are replaced by \texttt{<NUM>}.
	Similarly, mentions of names regarding the reviewed items are replaced by \texttt{<ORG>}.
	For each domain, \texttt{<UNK>} is used to alter out-of-vocabulary words (viz. terms that exist in the training set but are missing from validation/test set) in the reviews.
	
	Finally, the three types of context are assembled for individual reviews within each partition following Equation~\eqref{eq:context:neighbors}.
	For each domain, the constructed review-neighbors pairs across reviewed items are gathered.
	Helpful review-neighbors pairs are randomly sampled to have the same number as unhelpful ones and vice versa to avoid class imbalance.
	Throughout this work, NAP and all baseline models are trained on the training set, tuned on the validation set, and evaluated on the test set serving as unseen data in reality. 
	
	Table~\ref{tab:context:stats} demonstrates the simple descriptive statistics.
	As seen, reviews posted in ConsumerAffairs tend to be roughly twice lengthier than those in SiteJabber.
	
	\begin{table*}[width=.9\textwidth,cols=8,pos=h]
		\centering
		\caption{Descriptive statistics of the balanced doamins after pre-processing.}
		\label{tab:context:stats}
		\begin{tabular*}{\tblwidth}{@{} LLLLLLLL@{} }
			\toprule
			\multicolumn{2}{l}{Domain} & \#Reviews & \#Words & $\frac{\text{\#Words}}{\text{\#Reviews}}$ & \#Sentences & $\frac{\text{\#Sentences}}{\text{\#Reviews}}$ & $\frac{\text{\#Words}}{\text{\#Sentences}}$ \\ \midrule
			D1 & Dating & 4,054 & 359,369 & 88.65 & 27,035 & 6.67 & 12.91 \\
			D2 & Wedding Dresses & 5,294 & 456,602 & 86.25 & 36,909 & 6.97 & 12.67 \\
			D3 & Marketplace & 6,964 & 581,456 & 83.49 & 46,222 & 6.64 & 12.31 \\
			D4 & Car Insurance & 2,932 & 398,341 & 135.86 & 27,004 & 9.21 & 14.42 \\
			D5 & Travel Agencies & 8,156 & 1,168,941 & 143.32 & 78,408 & 9.61 & 14.67 \\
			D6 & Mortgages & 4,602 & 652,223 & 141.73 & 44,955 & 9.77 & 14.13 \\ 
			\bottomrule
		\end{tabular*}
	\end{table*}
	
	\subsection{Baseline Methods}
	\label{sec:context:baselines}
	
	The three types of neighbor-aware helpfulness prediction (i.e., preceding, following, and surrounding reviews) are compared with the independent counterpart.
	In NAP, independent helpfulness prediction is achieved by setting $\gamma=1$ in Equation \eqref{eq:context:combine}.
	NAP is also benchmarked against six state-of-the-art baselines modeling helpfulness beyond individual reviews.
	For simplicity, independent helpfulness prediction is henceforth denoted as \textbf{I} and the three types of neighbor-aware helpfulness prediction \textbf{I+P}, \textbf{I+F}, \textbf{I+S}, respectively.
	
	\begin{itemize}
		\item \textbf{I+ORD}:
		This baseline operationalizes three types of orders \cite{Alzate2018Exploring, Zhou2017Order}.
		The first type, denoted as $\textbf{I+ORD}_\text{D}$, is based on review dates.
		Let $R$ be reviews of the same product sorted from the latest to the oldest, each review $r \in R$ is associated with a posted date $d_r$.
		Given a day $d' \in \{d_r \mid r \in R\}$, reviews $R_{d'} \equiv \{r \mid d_r = d'\}$ posted on the same day are shared with the same order $\left[\sum_{d < d'} N(R_d) + 1\right]^{-1}$, where $N(R_d)$ is the cardinality of $R_d$.
		Similarly, the second type $\textbf{I+ORD}_\text{R}$ and third type $\textbf{I+ORD}_\text{V}$ of orders are handled respectively by sorting reviews from highest to lowest star ratings and from the largest to smallest number of helpful votes.
		\item \textbf{I+CON}: This baseline measures the conformity \cite{Lu2010} of a review $r \in R$ to reviews $R$ of the same product.
		Each review $r \in R$ is first vectorized into its unigram TFIDF representation $\mathbf{u}_r$.
		The conformity calculates the Kullback--Leibler divergence between a review $\mathbf{u}_r$ and the overall opinion $\bar{\mathbf{u}} = \frac{1}{|R|}\sum_{r \in R}{\mathbf{u}_r}$.
		\item \textbf{I+POL}: This baseline measures the sentiment divergence \cite{Hong2012} of a review $r \in R$ from reviews $R$ of the same product.
		Each review $r \in R$ is associated with (i) a numeric polarity $p_r \in [-1, 1]$ decided by the proportion of positive and negative words in $r$ and (ii) a categorical polarity $c_r \in \{\text{negative}, \text{neutral}, \text{positive}\}$ based on $p_r$.
		The divergence results from the absolute difference between a review $p_r$ and the mainstream opinion $\bar{p} = \frac{1}{|R'|}\sum_{r' \in R'} p_r'$, where $R' \equiv \{r' \mid c_r = c'\}$ and $c'$ belongs to the categorical polarity that shared by the majority of reviews in $R$.
		\item \textbf{I+ENT}: This baseline measures the incremental information entropy \cite{Jorge2019Entropy} of reviews $R$ of the same product. 
		Let $R_n$ be the $n$-th review, $n \in \mathbb{N}^{+}$, $\text{vocab}(R_n)$ returns the total number of unique words occurred in $\{R_m \mid m=1,2,\ldots,n\}$. 
		The entropy increment of $R_n$ is defined as $\text{vocab}(R_n) - \text{vocab}(R_{n-1})$, which computes the increased number of unique words in $R_n$ beyond that have been mentioned in $\{R_1, R_2, \ldots, R_{n-1}\}$.
	\end{itemize}
	
	In the baselines above, the proposed contextual information is used in conjunction with many other features, which are out of the scope of this work.
	To enable fair comparison, the orders extracted as per each baseline and the text embeddings $\mathbf{h}$ learned via \textbf{I} are concatenated and then fed into a feedforward layer for helpfulness prediction.
	NAP mainly differs from the baselines in that it locally takes neighbors of a review rather than the whole list of reviews as context.
	
	\subsection{Hyperparameters}
	\label{sec:context:hyper}
	
	The lookup table $\textbf{E}$ is initialized with the $300$-dimensional public-available GloVe word embeddings \cite{Pennington2014GloVe} and kept static during training.
	NAP employs $m=100$ kernels of patch size $l=3$ for review text encoding.
	Inspired by that most customers pay attention to no more than $10$ reviews \cite{Askalidis2016} before making purchase decisions, the number of neighbors $K$ for context construction is chosen between $1$ and $10$.
	The combination factor $\gamma$ is initially set to $0.5$ to assign equal importance to both the current review and its context.
	The weight decay for kernel regularization is set to $5 \times 10^{-4}$.
	
	The remaining network weights are initialized using the Glorot uniform initializer \cite{glorot2010} and updated through stochastic gradient descent over shuffled mini-batches of size $64$ using the Adam \cite{Diederik2014} update rule.
	During training, early stopping is applied when the validation loss has no improvement for $10$ epochs. 
	
	For reproducibility, all randomization processes involved in the experiments are initialized with the same random seed.
	The training of each model/baseline is repeated five times to test model robustness under different random initialization.
	
	\section{Result Analysis and Discussions}
	\label{sec:context:results}
	
	NAP is first quantitatively evaluated via extensive experiments, followed by discussions on the effectiveness of NAP and model sensitivity to different context settings.
	Qualitative analysis is then conducted.
	Throughout the experiments, model performance is measured by classification accuracy.
	
	\subsection{Comparison with Baseline Methods}
	
	Table~\ref{tab:context:results} benchmarks NAP against the baselines.
	The used context settings of \textbf{I+P}, \textbf{I+F}, and \textbf{I+S} are based on those yielding the highest performance.
	In the table, results outperforming both the independent counterpart \textbf{I} and the baselines are in italic, whereas the highest results are in bold.
	
	In brief, NAP achieves the highest accuracy across domains and leads by approximately $1\%$ to $5\%$.
	On average, NAP engages eight neighbors for context construction.
	In terms of weighting schemes, WAVG and FR are more frequently adopted than AVG and SFR.
	Section \ref{sec:context:consettings} will further investigate the context settings.
	In contrast, the six baselines are less robust to different domains.
	In the experiments, most improvements are observed on D1 and D2. On D3, D4, and D6, the introduced contextual features do not influence \textbf{I} or even diminish the performance.
	Over all domains, the contextual features yield less than $1\%$ accuracy gains.
	
	\begin{table*}[width=.85\textwidth,cols=12,pos=h]
		\caption[The results of NAP against the baseline methods.]{The results of NAP against the baseline methods. The context settings (Weighting Scheme/\#Neighbors) that produce the highest accuracy are listed below.}
		\label{tab:context:results}
		\begin{tabular*}{\tblwidth}{@{} LL|LLLLLL|LLL@{} }
			\toprule
			& \textbf{I} & $\textbf{I+ORD}_\text{D}$ & $\textbf{I+ORD}_\text{R}$ & $\textbf{I+ORD}_\text{V}$ & \textbf{I+CON} & \textbf{I+POL} & \textbf{I+ENT} & \textbf{I+P} & \textbf{I+F} & \textbf{I+S} \\
			\midrule
			D1 & 86.27 & 86.46 & 86.36 & 86.46 & 86.27 & 86.89 & 86.65 & \textit{89.90}$^{}$ & \textit{90.86}$^{}$ & \textbf{90.91}$^{}$ \\
			& & & & & & & & FR/9 & SFR/10 & WAVG/8 \\
			D2 & 70.04 & 70.20 & 70.90 & 70.16 & 70.47 & 70.66 & 70.66 & \textit{70.98}$^{}$ & \textit{71.17}$^{}$ & \textbf{71.25}$^{}$ \\
			& & & & & & & & FR/10 & AVG/7 & WAVG/10 \\
			D3 & 81.63 & 80.52 & 81.36 & 80.95 & 80.54 & 80.16 & 80.73 & \textit{83.56}$^{}$ & \textbf{83.83}$^{}$ & \textit{83.80}$^{}$ \\
			& & & & & & & & AVG/10 & FR/10 & FR/10 \\
			D4 & 72.21 & 71.64 & 72.05 & 71.39 & 72.13 & 71.56 & 72.38 & \textit{74.84}$^{}$ & \textit{74.59}$^{}$ & \textbf{75.00}$^{}$ \\
			& & & & & & & & SFR/6 & WAVG/3 & FR/6 \\
			D5 & 67.15 & 67.12 & 67.06 & 67.57 & 67.18 & 67.54 & 66.93 & \textit{67.96} & \textbf{68.41} & \textit{67.80} \\
			& & & & & & & & FR/10 & FR/7 & WAVG/8 \\
			D6 & 69.39 & 69.04 & 69.48 & 69.39 & 69.39 & 69.13 & 69.39 & \textbf{70.87}$^{}$ & \textit{70.43}$^{}$ & \textit{70.35}$^{}$ \\
			& & & & & & & & WAVG/10 & WAVG/6 & WAVG/4 \\ 
			\bottomrule
		\end{tabular*}
	\end{table*}
	
	\subsection{What Makes NAP Effective?}
	\label{sec:context:ca}
	
	In NAP, each review is contextualized within its neighbors for helpfulness prediction.
	Therefore, the performance gains of NAP compared with \textbf{I} and the baselines can result from (i) the interaction between a review and its neighbors, (ii) the exclusive context learned from the neighbors, or (iii) simply an increase of review data for model training.
	To validate the factors that lead to the effectiveness of NAP, the following NAP variants are evaluated:
	
	\begin{itemize}
		\item \textbf{P}/\textbf{F}/\textbf{S}:
		Neighbor-only prediction using merely the context embedding $\textbf{c}$ for helpfulness modeling, by setting $\gamma=0$ in Equation~\eqref{eq:context:combine}.
		The three types of neighbors: preceding reviews, following reviews, and surrounding reviews, are considered.
		\item \textbf{I+R}: 
		Neighbor-aware prediction where the context embedding $\mathbf{c}$ encodes the same number of $K$ reviews randomly selected from the same domain. 
		\item \textbf{I+N}: 
		Neighbor-aware prediction where the context embedding $\mathbf{c}$ draws random values from a uniform distribution within the range $[0, 1]$.
		This variant can also be thought of as introducing noise information into independent helpfulness prediction \textbf{I}.
	\end{itemize}
	
	\subsubsection{Neighbor-aware versus Neighbor-only}
	\label{sec:context:ca_context}
	
\begin{figure*}[htbp]
		\centering
		\includegraphics[width=0.75\textwidth]{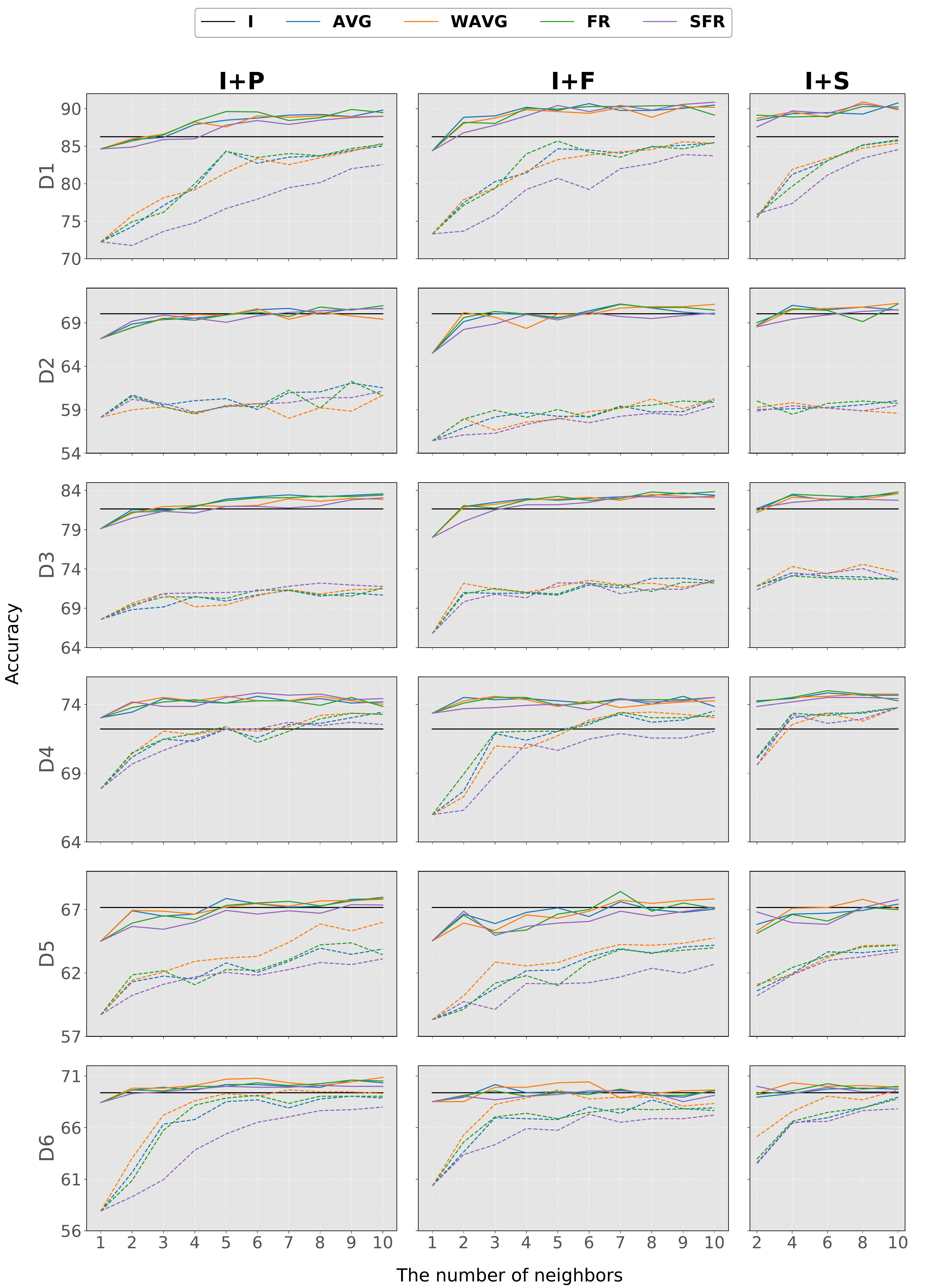}
		\caption[The performance of NAP on different context settings.]{The performance of NAP on different context settings. Dotted lines are the neighbor-only counterparts of the neighbor-aware methods. 
		}
		\label{fig:context:ca_ind}
	\end{figure*}

	Figure~\ref{fig:context:ca_ind} compares the neighbor-aware with neighbor-only methods to validate the role of neighbors during helpfulness prediction.
	As depicted, \textbf{P}, \textbf{F}, and \textbf{S} receive significantly lower performance than \textbf{I+P}, \textbf{I+F}, and \textbf{I+S} across domains, respectively.
	The only exception is D6 where $K=7$ following neighbors are weighted using the WAVG scheme, which produces less than $0.1\%$ increase in accuracy.
	The results strongly evidence that the effectiveness of NAP lies in an independent review interacting with its neighbors rather than either of the individuals.
	
	Overall, both the neighbor-aware and neighbor-only methods benefit from involving more neighbors.
	Recall that neighbors are treated as prior knowledge to support helpfulness interpretation.
	Involving more neighbors helps \textbf{P}, \textbf{F}, and \textbf{S} accumulate helpfulness clues, which may include those could have been mentioned in the targeted review.
	As a result, the accuracy of \textbf{P}, \textbf{F}, and \textbf{S} is gaining faster as $K$ increases and less likely to plateau. 
	Still, the accumulated clues can hardly cover all information contained in the targeted review.
	This explains why the neighbor-aware methods achieve higher accuracy than the neighbor-only counterparts with far fewer neighbors. 
	Without knowledge of the targeted review, \textbf{P}, \textbf{F}, and \textbf{S} also perform less stably across weighting schemes and neighbor types than the neighbor-aware methods.
	
	In several cases, the neighbor-only methods show comparable predictive power to independent helpfulness prediction.
	On D1, for instance, the accuracy of \textbf{P}, \textbf{F}, and \textbf{S} is close to \textbf{I} at $K = 10$.
	On D4, neighbor-only methods outperforming \textbf{I} is observed using $K \geq 4$ reviews.
	This suggests that the helpfulness of a review can sometimes be approximated by the collective helpfulness of its neighbors.
	On the majority of occasions, however, the effectiveness of the neighbor-only methods is weak.
	
	\textbf{Lessons Learned}: The performance gains of NAP mainly result from the review-neighbors interaction.
	Using neighbors alone, while comparable in rare cases, is not effective for helpfulness prediction. 
	
	\subsubsection{Neighbors versus Non-neighbors}
	\label{sec:context:ind_rnd}
	
	To validate whether the performance gains result from simply inputting more reviews, the neighbors used in NAP are replaced by the two types of non-neighbor context \textbf{I+N} and \textbf{I+R}.
	Note that the SFR weighting scheme is excluded from \textbf{I+R} since random reviews do not possess spatial characteristics.
	
\begin{figure}[htbp]
		\centering
		\includegraphics[width=1\linewidth]{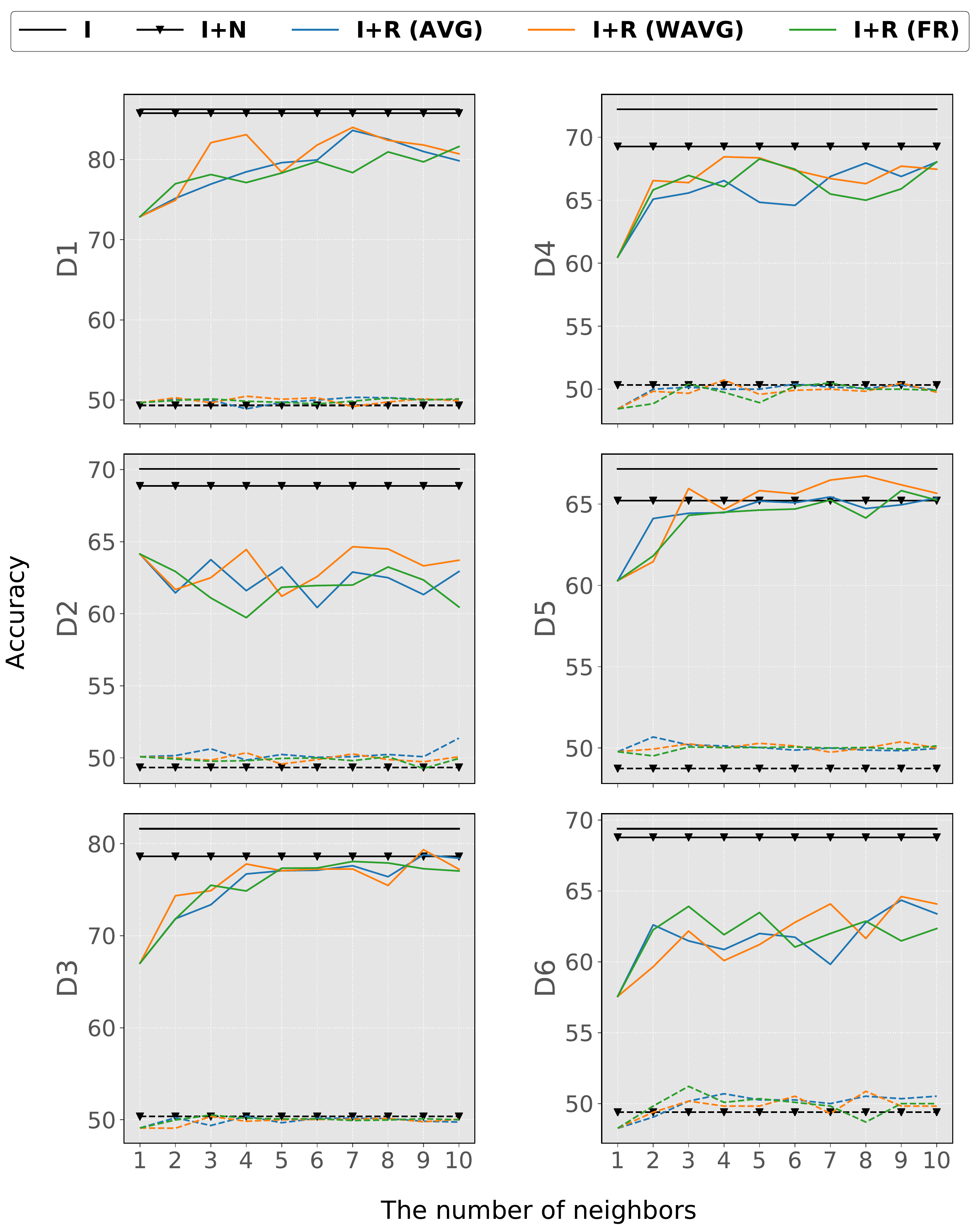}
		\caption[The performance of NAP using non-neighbor context.]{The performance of NAP using non-neighbor context. Dotted lines are the context-only counterparts.}
		\label{fig:context:rnd_nei}
	\end{figure}

	As shown in Figure~\ref{fig:context:rnd_nei}, both types of non-neighbor context receive lower accuracy than \textbf{I+P}, \textbf{I+F}, \textbf{I+S}, and \textbf{I} across domains. 
	Similar to \textbf{I+N}, \textbf{I+R}
	can be thought of as introducing a form of noise into \textbf{I}.
	Although involving more random reviews tends to improve \textbf{I+R}, the accuracy across domains, if not comparable to, is worse than \textbf{I+N}.
	This suggests that random reviews \textbf{R} harm \textbf{I} even more than random noise \textbf{N}.
	On the other hand, using \textbf{N} alone acts similarly to random guessing ($50\% \pm 2.5\%$).
	The performance of \textbf{R} fluctuates around \textbf{N} regardless of the value of $K$. 
	Compared with \textbf{P}, \textbf{F}, and \textbf{S}, simply stacking random reviews cannot accumulate helpfulness clues to form an effective context.
	The results prove the indispensability of using neighbors for context construction.
	
	\textbf{Lessons Learned}: 
	The effectiveness of NAP essentially relies on learning specific clues from neighbors.
	Simply including arbitrary reviews does not lead to performance gains.
	
	\subsection{Sensibility Analysis on Context Settings}
	\label{sec:context:consettings}
	
	Four types of NAP hyperparameters are further explored to investigate how different context settings affect the model.
	The hyperparameters and their possible values are listed in Table~\ref{tab:context:hyper}.
	Subsequently, the trade-off between NAP's performance and complexity is discussed.
	
	\begin{table}[width=0.95\linewidth,cols=2,pos=h]
		\centering
		\caption{NAP context settings to be investigated.}
		\label{tab:context:hyper}
		\begin{tabular*}{\tblwidth}{@{} p{0.5\linewidth}p{0.4\linewidth}@{} }
			\toprule
			Hyperparameters & Possible Values \\
			\midrule
			The number of neighbors $K$ & $\{i \mid i \in \mathbb{N}^+, 1 \leq i \leq 10\}$ \\
			The neighbor selection schemes & Previous, following, and surrounding neighbors \\
			The weighting schemes & AVG, WAVG, FR, SFR \\
			The combination factor $\gamma$ & $\{\frac{i}{10} \mid i \in \mathbb{N}^+, 1 < i < 10\}$ \\
			\bottomrule
		\end{tabular*}
	\end{table}
	
	\subsubsection{Number of Neighbors}
	\label{sec:context:ana_k}
	
	Figure~\ref{fig:context:ca_ind} illustrates the relationship between the number of neighbors and model performance.
	As shown, NAP generally improves as $K$ increases and then plateaus.
	Most domains reach the highest accuracy with a $K$ value close to $10$, but the performance gains after the first few neighbors are less than $1.5$\%.
	This confirms that neighbors closer to a review drive the bulk of the influence on customers perceiving review helpfulness.
	Taking all neighbor types and weighting schemes into account, NAP initially beats \textbf{I} within the first five reviews.
	In particular, all domains but D5 achieve so within only the first two reviews. 
	
	Overall, NAP is inferior to \textbf{I} 
	when learning context from extremely few neighbors.
	In a way analogous to \textbf{I+N}, the insufficient context information used in NAP can be thought of as introducing noise to \textbf{I}.
	NAP starts to improve and outperform \textbf{I} when more neighbors are involved.
	The additional neighbors aid consolidating contextualization by accumulating helpfulness clues.
	At some point, continuing to include neighbors has little influence on NAP, suggesting that the information needed for contextualization has saturated.
	
	\subsubsection{Neighbor Selection Schemes}
	
\begin{figure}[htbp]
		\centering
		\includegraphics[width=0.71\linewidth]{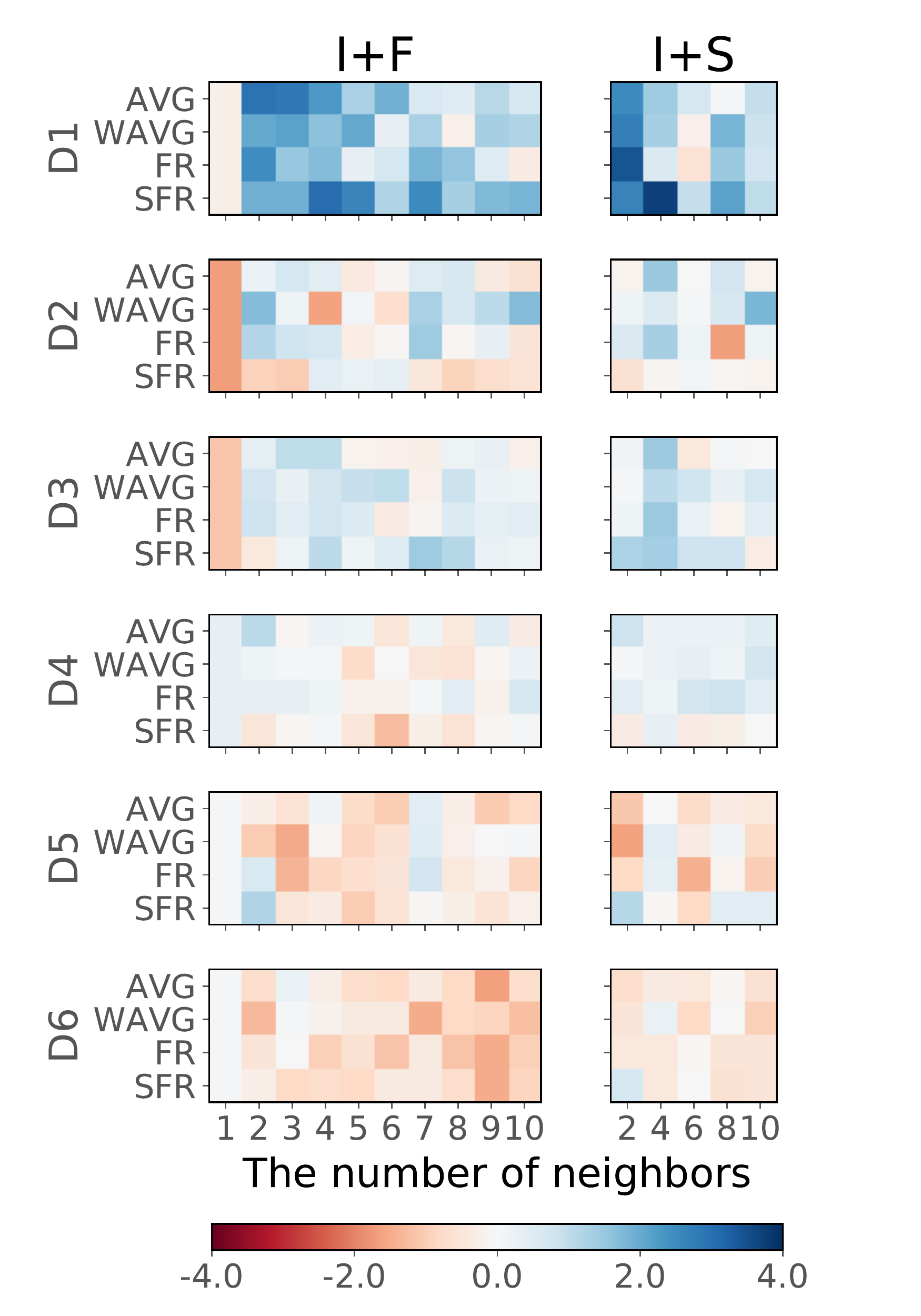}
		\caption[Performance comparison among the three neighbor types.]{The increase/decrease in accuracy of \textbf{I+F} and \textbf{I+S} compared with \textbf{I+P}.}
		\label{fig:context:preceding_based}
	\end{figure}

	The three neighbor selection schemes are compared.
	In particular, \textbf{I+P} is selected as the baseline to observe the change of performance from using preceding neighbors to following and surrounding ones as context.
	Figure~\ref{fig:context:preceding_based} demonstrates the domain-dependent behavior of neighbor selection.
	On D1--D4, \textbf{I+P} generally outperform \textbf{I+F} and \textbf{I+S}, suggesting that customers rely more on preceding neighbors to determine review helpfulness. 
	Contrarily, following and surrounding reviews are more capable on D5 and D6 of constructing context information.
	The performance gaps among the neighbor types are mostly within $2$\%.
	
	\subsubsection{Weighting Schemes}
\begin{figure}[htbp]
		\centering
		\includegraphics[width=1\linewidth]{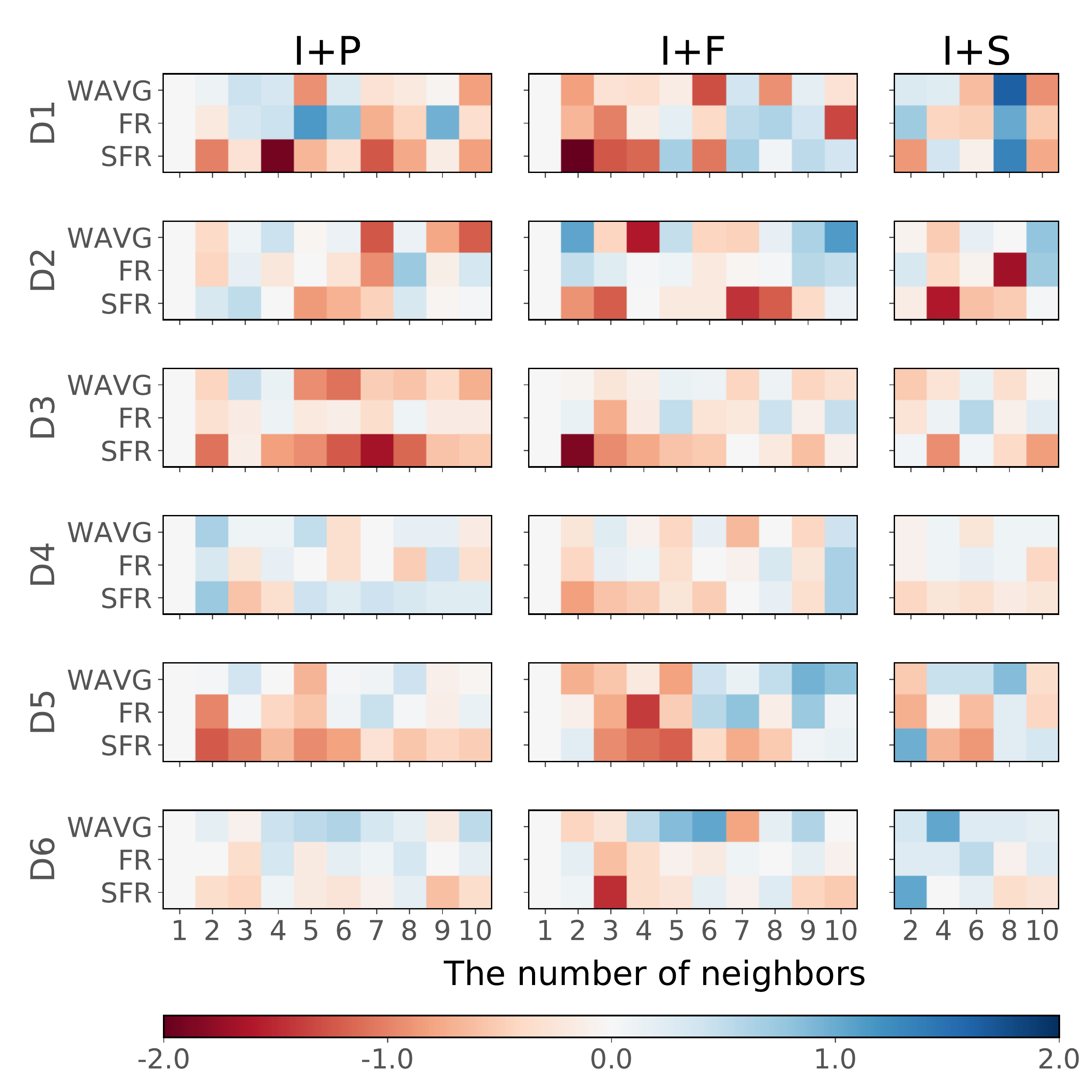}
		\caption[Performance comparison among the four weighting schemes.]{The increase/decrease in accuracy of other weighting schemes compared with AVG.}
		\label{fig:context:avg_based}
	\end{figure}

	In a similar vein, Figure~\ref{fig:context:avg_based} compares the four weighting schemes by computing the performance gaps between AVG and the rest.
	As shown, AVG offers a robust option for learning context clues from neighbors, with the gap within $1$\% ($2$\%) in most (all) cases.
	The highest performance (blocks in the darkest blue colors) is majorly achieved by either WAVG or FR, necessitating the use of finer-grained schemes to gather useful information from neighbors of uneven quality.
	Whereas modeling neighbor interactions during context construction brings less obvious improvement.
	In many cases, SFR receives lower accuracy than other schemes if not having comparable performance.
	This requires further analysis on the interaction mechanism among neighbors in future work.
	
	\subsubsection{Combination Factor}
	
\begin{figure}[htbp]
		\centering
		\includegraphics[width=1\linewidth]{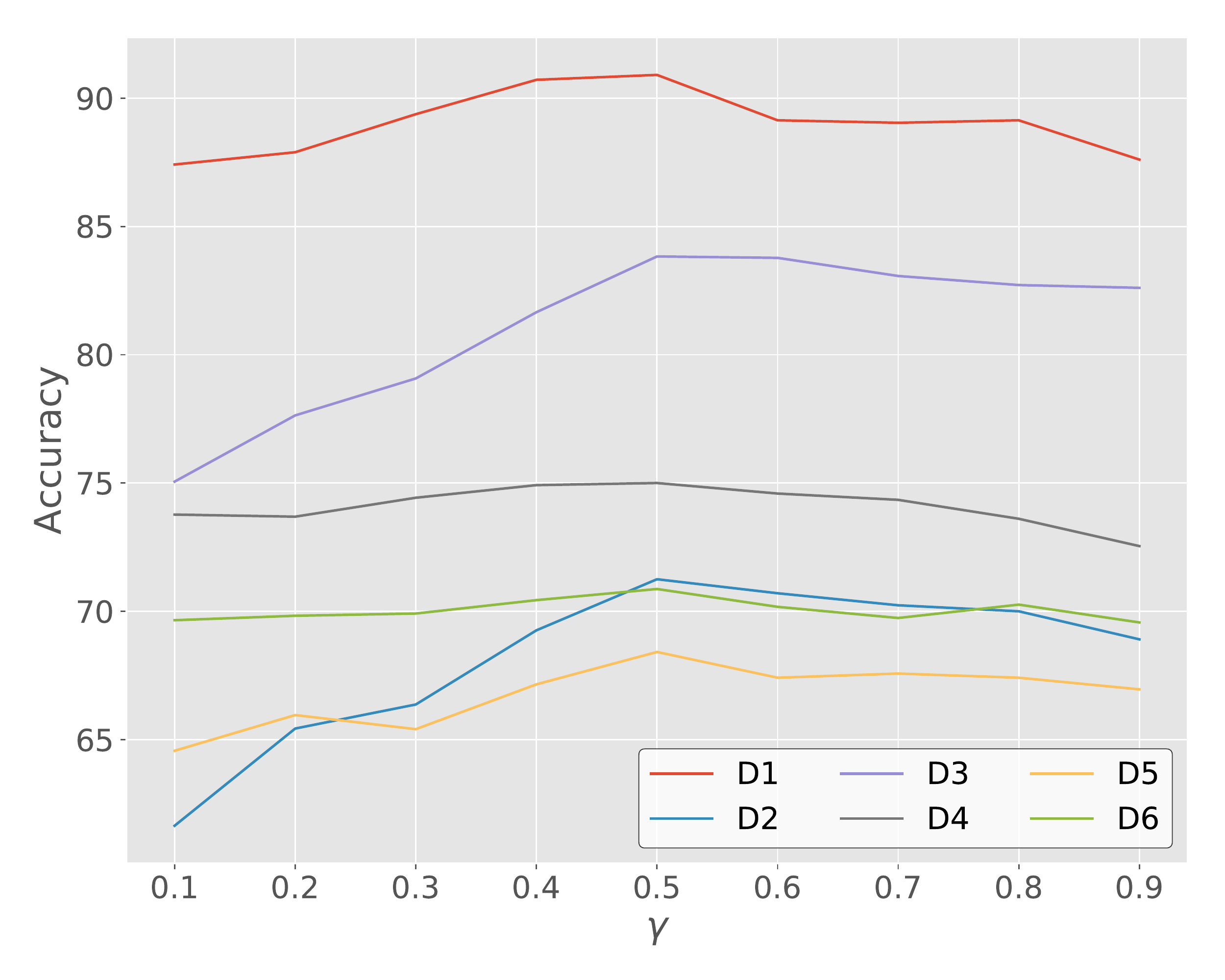}
		\caption[The performance of NAP on different $\gamma$ values.]{The performance of NAP on different $\gamma$ values. From left to right, the influence of neighbors on a review decreases. 
		}
		\label{fig:context:gamma}
	\end{figure}

	Figure~\ref{fig:context:gamma} analyzes the combination factor $\gamma$ controlling the influence of neighbors on a current review during contextualization.
	The value of $\gamma$ is varied from $0.1$ to $0.9$ incremented by $0.1$, using the context settings mentioned in Table~\ref{tab:context:results}.
	The two cases $\gamma=0$ (i.e., neighbor-only helpfulness prediction) and $\gamma=1$ (i.e., independent helpfulness prediction) are ignored as has been reported in previous sections.
	Recall in Equation~\eqref{eq:context:combine} that the value of $\gamma$ is inversely proportional to the influence of neighbors.
	
	As shown, the sensitivity of NAP to $\gamma$ differs across domains.
	Overall, the performance of NAP first increases and then decreases along with $\gamma$, peaking at around $\gamma=0.5$.
	This suggests neither excessive dependence on a current review or that on its neighbors facilitates contextualized helpfulness prediction. 
	The finding further confirms that the effectiveness of NAP results from the review-neighbor interaction rather than only either source.
	
	While acting similarly across domains in $\gamma \in [0.5, 0.9]$, NAP is more sensitive to the amount of neighbor information used for helpfulness modeling in $\gamma \in [0.1, 0.5]$.
	Specifically, D2 and D3 show relatively high sensitivity, followed by D1 and D5, and finally D4 and D6 are comparatively less sensitive to $\gamma$.
	One explanation is the difference in domain-specific characteristics, for instance, the homogeneity of review opinions towards the same product.
	
	\subsubsection{Trade-off between Performance and Complexity}
	
	As has been shown in Table~\ref{tab:context:results}, NAP tends to involve large number of neighbors and more flexible weighting schemes.
	Although reaching the highest performance, such context settings demand high computational complexity.
	Table~\ref{tab:context:complexity} summarizes the number of floating point operations and that of parameters required during context construction.
	As discussed, using excessive number of neighbors and/or overcomplicated weighting schemes does not guarantee a significant increase of accuracy.
	In circumstances where efficiency is emphasized, the relatively slight improvement can be traded for a faster model implementation.
	
	\begin{table}[width=0.9\linewidth,cols=3,pos=h]
		\centering
		\caption{NAP complexity on different weighting schemes.}
		\label{tab:context:complexity}
		\begin{tabular*}{\tblwidth}{@{} LLL@{} }
			\toprule
			Scheme & Floating Point Operations$^*$ & Parameters \\
			\midrule
			AVG & $mK$ & $0$ \\
			WAVG & $2mK+3K+1$ & $m$ \\
			FR & $5mK+2m$ & $mK$ \\
			SFR & $(mK^2+11mK)/2+2m$ & $mK$ \\
			\bottomrule
			\multicolumn{3}{l}{* The number of operations at each epoch.} \\
			\multicolumn{3}{l}{Bias terms are omitted for simplicity.} \\
		\end{tabular*}
	\end{table}
	
	This section searches for alternative NAP context settings that reduce model complexity while maintaining performance within an acceptable range.
	Let $p$ be the context setting in a domain that leads to the highest accuracy $q$, $\hat{p}$ is a comparable alternative for $p$ if (1) $\hat{p}$ uses smaller $K$ values, (2) $\hat{p}$ uses simpler weighting schemes, and (3) $|\hat{q} - q| \leq \delta$.
	Here, $\delta \in [0, 1]$ constrains the drop of performance to be no more than $1$\%.
	Table~\ref{tab:context:suboptimals} lists the alternative context settings ordered by $\delta$.
	As shown, comparable neighbor-aware helpfulness prediction can be approached using AVG on at most five neighbors, with less than $0.72$\% accuracy drop.
	Among these alternative settings, following and surrounding reviews tend to be more effective neighbor selection schemes.
	
	\begin{table}[width=0.9\linewidth,cols=5,pos=h]
		\centering
		\caption{Alternative context settings.}
		\label{tab:context:suboptimals}
		\begin{tabular*}{\tblwidth}{@{} LLLLL@{} }
			\toprule
			& \makecell{Weighting \\ Scheme} & \makecell{Neighbor \\ Scheme} & $K$ & $\delta$ \\ \midrule
			D1 & AVG & \textbf{I+F} & 6 & 0.2392 \\
			& AVG & \textbf{I+F} & 4 & 0.7177 \\ \midrule
			D2 & AVG & \textbf{I+F} & 7 & 0.0781 \\
			& AVG & \textbf{I+S} & 4 & 0.2344 \\ \midrule
			D3 & FR & \textbf{I+F} & 8 & 0.0272 \\
			& FR & \textbf{I+S} & 4 & 0.3261 \\
			& AVG & \textbf{I+S} & 4 & 0.4348 \\ \midrule
			D4 & AVG & \textbf{I+S} & 6 & 0.1639 \\
			& AVG & \textbf{I+F} & 2 & 0.4918 \\ \midrule
			D5 & AVG & \textbf{I+P} & 5 & 0.5502 \\ \midrule
			D6 & WAVG & \textbf{I+P} & 6 & 0.0870 \\
			& WAVG & \textbf{I+P} & 5 & 0.1739 \\
			& WAVG & \textbf{I+S} & 4 & 0.5217 \\
			& AVG & \textbf{I+F} & 3 & 0.6957 \\ \bottomrule
		\end{tabular*}
	\end{table}
	
	\subsection{Qualitative Analysis}
	\label{sec:context:qualana}
	
	Two qualitatively analysis tasks are conducted to provide more straightforward and explainable evidence towards the effectiveness of NAP.
	As an example, D1 using the first alternative context setting (
	averaging the opinions of six following neighbors of a current review
	) in Table~\ref{tab:context:suboptimals} is selected.
	
	\subsubsection{Learned Document Embeddings}
\begin{figure}[htbp]
		\centering
		\includegraphics[width=0.5\textwidth]{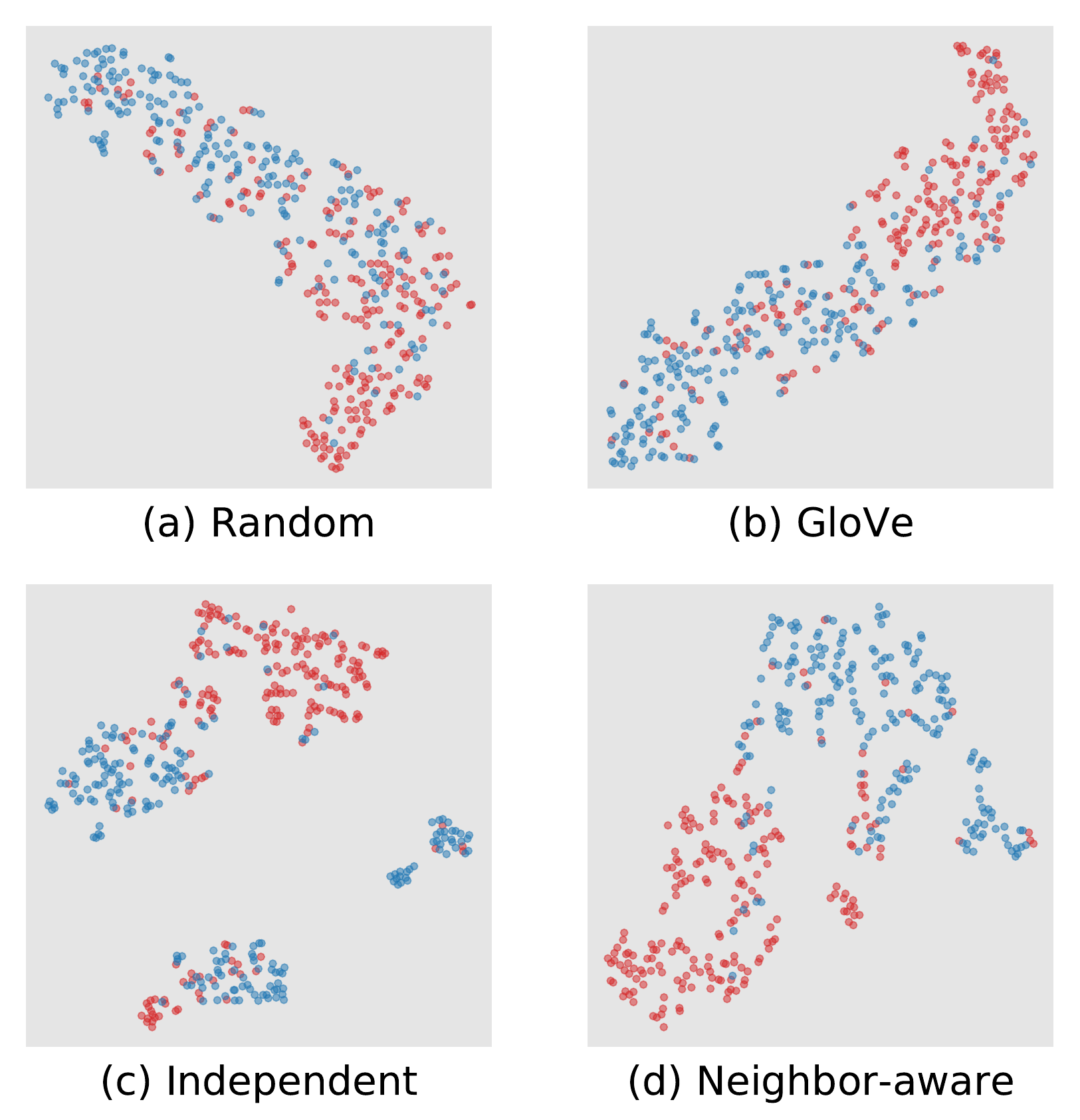}
		\caption[$t$-SNE projection of the learned document embeddings.]{$t$-SNE projection of the learned document embeddings.
			Blue and red points are helpful and unhelpful reviews, respectively. 
			(a) The model weights are initialized randomly. (b) Similar to (a) except the embedding table $\textbf{E}$ is initialized by pre-trained GloVe embeddings. (c) The weights are trained for independent helpfulness prediction. (d) The weights are trained for neighbor-aware helpfulness prediction.}
		\label{fig:context:doc_emb}
	\end{figure}

	The first task illustrates the learned neighbor-aware document embeddings of testing samples produced by NAP for helpfulness prediction.
	To this end, the output of the penultimate layer (Equation~\eqref{eq:context:combine}) is computed.
	As for dimensionality reduction, $t$-SNE \cite{Maaten2008visualizing} is applied to obtain the corresponding $2$-dimensional vector representations.
	Figure~\ref{fig:context:doc_emb} presents the predicted document embeddings using neural network weights before and after model training.
	When the weights are initialized randomly, helpful and unhelpful samples are mixed with each other.
	Replacing the random weights in the embedding table $\textbf{E}$ with those pre-trained by GloVe does not lead to significant difference.
	When NAP is trained, the weights of both independent and neighbor-aware helpfulness prediction can effectively separate helpful and unhelpful samples.
	In particular, the latter learn better separability to distinguish helpful reviews from unhelpful ones.
	Therefore, the use of review neighbors as context strengthens the predictive power of helpfulness prediction.
	
	\subsubsection{Case Studies}
	The second task investigates possible reasons of a current review being influenced by its neighbors in reality.
	Table~\ref{tab:context:showcasing} provides four instances from the test set, each containing a review and its neighbors, along with the predicted helpfulness and ground-truth labels.
	In (a), the helpfulness of the current review per se is ambiguous.
	Given the context mostly mentioning a similar issue of insufficient member interaction (as underlined in the table), the current review is more trustworthy and thus wins additional helpfulness.
	Similarly, the neighbors in (b) aid forming an impression that the dating platform mainly suffers from pricing and customer service.
	Such context confirms and supports the current review, making it more helpful than it could have been if presented alone.
	On the contrary, (c) and (d) show a different scenario where the formed impression (overall positive) contradicts the current review's opinion (overall negative).
	In this case, the context weakens the perceived standalone helpfulness of the current review.
	The four instances above show that neighbor-aware helpfulness prediction can surpass its independent counterpart by capturing the influence brought by review neighbors.
	
	\begin{table*}[width=1\textwidth,cols=3,pos=htbp]
		\centering
		\caption[Examples of real-world reviews influenced by their neighbors.]{Examples of real-world reviews influenced by their neighbors. 
			Each example contains six reviews as the context of the current review.
			From left to right, each helpfulness triplet indicates (1) the predicted independent helpfulness, (2) the predicted neighbor-aware helpfulness, and (3) the ground-truth label.
		}
		\label{tab:context:showcasing}
		\footnotesize
		\begin{tabular*}{\tblwidth}{@{} lp{0.83\textwidth}p{0.1\textwidth}@{} }
			\toprule
			& Review & Helpfulness \\ 
			\midrule
			\multirow{2}{*}{(a)} & 
			\begin{enumerate}[noitemsep,topsep=0pt]
				\item ``I was disappointed and they took my money but \underline{no dates after 6 months} of subscriptions. [...] [N]o one sent me any email or respond, \underline{no connections or dates}. [...]''
				\item ``Sorry I joined. I joined a few weeks ago. I have seen \underline{no new people since then}. The site is often down. I am not pleased and wish I had not first joined for a year.''
				\item ``Horrible Experience. The screening process never produced results. [...] This is poor customer service and hiding behind policy when a customer is unhappy says a lot about their poor product.''
				\item ``\texttt{<ORG>} [w]as the worst experience I've ever had!!! [...] [K]ept charging my credit card and \underline{never gave me any dates}! It's a bunch of young kids running the office and don't have a clue what they are doing!''
				\item ``I found a way to close my \texttt{<ORG>} account. I joined this dating site 4 weeks ago and didn't like the fact \underline{I wasn't being matched} with the women I put in my profile [...]''
				\item ``\underline{Not for those looking for real people to date}. [...] [Y]et I really have had \underline{no success with any matches}. It is very disappointing to be matched with at least 7 guys every day and get no response from any of them. [...]''
			\end{enumerate} 
			& 0--1--1
			\\
			& 
			``Senior dating? I signed up my dad to \texttt{<ORG>} to see if this site works for seniors and apparently it doesn't (at least not for him due to \underline{lack of members from his town}).''
			&  \\ \midrule
			\multirow{2}{*}{(b)} & 
			\begin{enumerate}[noitemsep,topsep=0pt]
				\item ``Canceled account, Still charged full price. I wish I would have read more reviews before agreeing to try \texttt{<ORG>}. [...] Something is not right about that. :-(''
				\item ``\texttt{<ORG>} - Rip OFF. [...] [E]ven when you deactivate your account, \texttt{<ORG>} will still \underline{charge your card}. [...] [N]avigating the site and working with \underline{`customer service' staff is a nightmare}! Never again!!!''
				\item ``Don't Do It... Unreliable and unethical. Their \underline{customer service is horrible}, \underline{the site itself is not user friendly} [...] My card \underline{was charged again} after having my account deactivated and their excuse was [...]''
				\item ``They will take your money. Be careful!!! Once you inactivate your account you will \underline{continue to get charged}. \texttt{<ORG>} will refund only one of these charges as a `courtesy'. [...] \underline{Customer service is completely non-existent}. [...]''
				\item ``Stay away. Hackers and scanners have hit this site, \texttt{<ORG>} needs ti tighten up their security. [...] I found \underline{no customer svc support} phone nimbers anywhwre on the site page.''
				\item ``Delporable Business Practice. [...]I called to request a credit as my profile had been taken down and I thought I had terminated my account on the site. Stupid me, I thought they would deal fairly with me. [...]''
			\end{enumerate}
			& 0--1--1
			\\
			& 
			``\underline{Glitches, cumbersome site and charged full price}!!!! I stupidly signed up for a year. I have met someone off line and haven't even been on the site a month. I have to pay for the entire year. Stay away from this site. [...] Horrible, horrible service!!!!!!!''
			&  \\ \midrule
			\multirow{2}{*}{(c)} & 
			\begin{enumerate}[noitemsep,topsep=0pt]
				\item ``Met an awesome lady. Thanks.''
				\item ``not too bad. better than the rest. not a hook up site for the most part.''
				\item ``Yes and No. [...] I like the offering of options to search that \texttt{<ORG>} gave me. I enjoyed the formatting and the contact options. [...] I did meet someone. Thank you for a wonderful experience.''
				\item ``God blessed me through \texttt{<ORG>}. [...] He is AMAZING and I truly feel God blessed me with this wonderful man. It is so good to have someone put such a smile on your face every day. I am one lucky lady!!''
				\item ``Met after 1 week. WE both joined about the same time. In a week we met, another week a 1st date, 5 weeks later am getting off on0line dating....hopefully for good.''
				\item ``Met the `Love of my Life' Great site that enabled us to meet and fall for each other!''
			\end{enumerate}
			& 1--0--0 
			\\
			& 
			``Awefull. I'm embarrassed that I got on \texttt{<ORG>}, I should have known better. The website is awe full to maneuver. [...] I recommended NOT TO SIGN UP ON THIS WEBSITE (for your own good and \$)''
			&  
			\\ \midrule
			\multirow{2}{*}{(d)} & 
			\begin{enumerate}[noitemsep,topsep=0pt]
				\item ``I met an awesome man on \texttt{<ORG>}. I wasn't going to join but this handsome man kept sending me messages and I had to see what he was saying. I'm so glad I did.''
				\item ``Located My Prince. [...] After a couple weeks of messaging, we began texting and talking on the phone. [...] We are now in a committed relationship and will be vacationing together this summer!''
				\item ``Hade a great time. Great site had a good experiences.''
				\item ``It may take time but someone is there for you. Don't give up. There are many good people on this site. I have actually met a few great guys.''
				\item ``its ok. same story as any site.''
				\item ``Finding love quickly. This is a wonderful site-found someone with in a week.''
			\end{enumerate}
			& 1--0--0
			\\
			& 
			``Total Rip Off. No matter how many miles you put, they keep sending you matches hundreds of miles away. People you contact are no longer on there. Once you cancel they use you profile forever. [...]''
			& 
			\\ \bottomrule
		\end{tabular*}
	\end{table*}
	
	\section{Conclusions and Future Work}
	\label{sec:context:con}
	
	This paper has proposed NAP for neighbor-aware helpfulness prediction.
	NAP differs from most existing studies that assume the perceived helpfulness of a review is self-contained.
	NAP also differs from existing context-aware methods that learn global context from a whole sequence of reviews.
	Instead, NAP contextualizes a review into a small number of its sequential neighbors, which better describes the reality.
	In NAP, a total of $12$ methods ($3$ neighbor selection schemes $\times$ $4$ weighting schemes) were explored for context construction from neighbors.
	Extensive experiments on six domains of real-world reviews were conducted to validate the feasibility and effectiveness of NAP.
	Empirical results and qualitative analysis show that exploiting the interaction between a review and its neighbors can improve helpfulness prediction and advance the state-of-the-arts.
	
	NAP was investigated under different context settings.
	Those producing the highest performance revealed that NAP engaged on average eight neighbors for context construction and considered the neighbors to be of uneven importance.
	The bulk of NAP's performance gains occurred in closer neighbors, whereas more distant ones had less influence.
	Selecting a type of neighbors for context construction, however, was domain-dependent, with 
	following and surrounding neighbors being more favoured
	.
	Cross-domain analysis further revealed that a highest-performance context setting could be approximated by averaging the opinions from no more than five closest neighbors of a review.
	The findings of this work will hopefully pave the way for future research in neighbor-aware helpfulness prediction.
	
	There are several directions to be addressed. 
	In the text encoding phase, more sophisticated representation methods will be employed to learn deeper semantics from review texts.
	As for context construction, more flexible schemes will be explored to select and aggregate neighbors.
	One example will be using skipped neighbors or asymmetrical surrounding neighbors. 
	In addition, a learned rather than specified combination factor can further automate the helpfulness modeling process.
	Finally, further analysis on NAP will be conducted to investigate the performance gaps among domains.
	It is also interested to check how including even more neighbors (e.g., up to $20$) will affect the performance of neighbor-aware helpfulness prediction.
	
	\bibliographystyle{cas-model2-names}
	
	\bibliography{Manuscript}
	
\end{document}